\newif\ifconfver
\confverfalse      
\confvertrue        

\newif\ifplainver  
\plainvertrue

\ifplainver
    \confverfalse   
\fi

\ifconfver
     \documentclass[10pt,twocolumn,twoside]{IEEEtran}
\else
    \ifplainver
        \documentclass[11pt]{article}
        \usepackage{fullpage}
    \else
        \documentclass[12pt,draftcls,onecolumn]{IEEEtran}
    \fi
\fi

\usepackage{booktabs,multirow}
\usepackage{calc,amsfonts,amssymb,amsmath,bm,url,color,theorem,graphicx,cite}
\usepackage{subfigure,float}
\usepackage{algorithm}
\usepackage{algorithmic}
\usepackage{enumerate}

\definecolor{orange}{RGB}{255,107,0}

\newtheorem{Fact}{Fact}

\newtheorem{Prop}{Proposition}
\newtheorem{Theorem}{Theorem}

\newtheorem{Corollary}{Corollary}

\theorembodyfont{\rmfamily}

\newcommand\bw{\ensuremath{{\bm w}}}
\newcommand\bW{\ensuremath{{\bm W}}}

\newcommand\bq{\ensuremath{{\bm q}}}
\newcommand\bx{\ensuremath{{\bm x}}}
\newcommand\by{\ensuremath{{\bm y}}}
\newcommand\bz{\ensuremath{{\bm z}}}

\newcommand\be{\ensuremath{{\bm e}}}

\newcommand\bX{\ensuremath{{\bm X}}}

\newcommand\bC{\ensuremath{{\bm C}}}
\newcommand\ba{\ensuremath{{\bm a}}}

\newcommand\bA{\ensuremath{{\bm A}}}

\newcommand\bb{\ensuremath{{\bm b}}}
\newcommand\bc{\ensuremath{{\bm c}}}
\newcommand\bd{\ensuremath{{\bm d}}}
\newcommand\bg{\ensuremath{{\bm g}}}

\newcommand\balp{\ensuremath{{\bm \alpha}}}

\newcommand\bF{\ensuremath{{\bm F}}}
\newcommand\bff{\ensuremath{{\bm f}}}

\newcommand\bD{\ensuremath{{\bm D}}}
\newcommand\bu{\ensuremath{{\bm u}}}

\newcommand\bPhi{\ensuremath{{\bm \Phi}}}

\newcommand\bV{\ensuremath{{\bm V}}}
\newcommand\bU{\ensuremath{{\bm U}}}

\newcommand\bs{\ensuremath{{\bm s}}}
\newcommand\bS{\ensuremath{{\bm S}}}

\newcommand{\Rbb}{\mathbb{R}}

\newcommand{\Sbb}{\mathbb{S}}

\newcommand{\setA}{\mathcal{A}}
\newcommand{\setB}{\mathcal{B}}
\newcommand{\setE}{\mathcal{E}}
\newcommand{\setF}{\mathcal{F}}
\newcommand{\setX}{\mathcal{X}}
\newcommand{\setR}{\mathcal{R}}

\newcommand{\setS}{\mathcal{S}}
\newcommand{\setT}{\mathcal{T}}
\newcommand{\setC}{\mathcal{C}}

\newcommand\bLam{\ensuremath{{\bm \Lambda}}}

\newcommand{\rbd}{{\rm rbd}}
\newcommand{\ri}{{\rm ri}}
\newcommand{\bdy}{{\rm bd}}
\newcommand{\inte}{{\rm int}}
\newcommand{\vol}{{\rm vol}}
\newcommand{\conv}{{\rm conv}}
\newcommand{\aff}{{\rm aff}}

\newcommand{\bzero}{{\bm 0}}
\newcommand{\bone}{{\bm 1}}
\newcommand{\bI}{{\bm I}}

\begin{document}

\bibliographystyle{IEEEtran}

\begin{center}
{\LARGE
Maximum Volume Inscribed Ellipsoid: A New Simplex-Structured Matrix Factorization Framework via Facet Enumeration and Convex Optimization} \\
~ \\
Chia-Hsiang Lin$^\dag$,
Ruiyuan Wu$^\ddagger$,
Wing-Kin Ma$^\ddagger$,
Chong-Yung Chi$^\S$, and
Yue Wang$^\star$
%
%
\\ ~ \\
$^\dag$Instituto de Telecomunica\c{c}\~{o}es, Instituto Superior T\'{e}cnico, Universidade de Lisboa, Lisbon, Portugal
\\
Emails:
\url{chiahsiang.steven.lin@gmail.com}
\\ ~ \\
$^\ddagger$Department of Electronic Engineering, The Chinese University of Hong Kong, Shatin, New Territories, Hong Kong.
\\
Emails:
\url{rywu@ee.cuhk.edu.hk},
\url{wkma@ieee.org}
\\ ~ \\
$^\S$Institute of Communications Engineering, National Tsing-Hua University, Hsinchu, Taiwan 30013, R.O.C.
\\
Emails:
\url{cychi@ee.nthu.edu.tw}
\\ ~ \\
$^\star$Department of Electrical and Computer Engineering, Virginia Polytechnic Institute and State University, VA, USA.
\\
Email:
\url{yuewang@vt.edu}
%
\\ ~ \\
\today
\end{center}


\newcommand{\paperabstract}{Consider a structured matrix factorization model where one factor is restricted to have its columns lying in the unit simplex.
This simplex-structured matrix factorization (SSMF) model and the associated factorization techniques have spurred much interest in research topics over different areas, such as hyperspectral unmixing in remote sensing, topic discovery in machine learning, to name a few.
In this paper we develop a new theoretical SSMF framework whose idea is to study a maximum volume ellipsoid inscribed in the convex hull of the data points.
This maximum volume inscribed ellipsoid (MVIE) idea has not been attempted in prior literature, and
we show a sufficient condition under which the MVIE framework guarantees exact recovery of the factors.
The sufficient recovery condition we show for MVIE is much more relaxed than that of separable non-negative matrix factorization (or pure-pixel search);
coincidentally it is also identical to that of minimum volume enclosing simplex, which is known to be a powerful SSMF framework for non-separable problem instances.
We also show that MVIE can be practically implemented by performing facet enumeration and then by solving a convex optimization problem.
The potential of the MVIE framework is illustrated by numerical results.

\bigskip

\noindent
{\bf Index Terms:}
maximum volume inscribed ellipsoid, simplex, structured matrix factorization, facet enumeration, convex optimization
}


\ifplainver


    \title{\papertitle}

    \author{
    Chia-Hsiang Lin
    }


    \begin{abstract}
    \paperabstract
    \end{abstract}

\else
    \title{\papertitle}

    \ifconfver \else {\linespread{1.1} \rm \fi

    \author{Chia-Hsiang Lin}

    \maketitle

    \ifconfver \else
        \begin{center} \vspace*{-2\baselineskip}
        \end{center}
    \fi

    \begin{abstract}
    \paperabstract
    \\\\
    \end{abstract}


    \ifconfver \else \IEEEpeerreviewmaketitle} \fi

 \fi

\ifconfver \else
    \ifplainver \else
        \newpage
\fi \fi

\section{Introduction}

Consider the following problem.
Let $\bX \in \Rbb^{M \times L}$ be a given data matrix.
The data matrix $\bX$ adheres to a low-rank model
$\bX = \bA \bS$, where $\bA \in \Rbb^{M \times N}, \bS \in \Rbb^{N \times L}$ with $N  \leq \min\{M,L\}$.
The goal is to recover $\bA$ and $\bS$ from $\bX$, with the aid of some known or hypothesized structures with $\bA$ and/or $\bS$.
Such a problem is called {\em structured matrix factorization (SMF)}.
In this paper we focus on a specific type of SMF called {\em simplex-SMF (SSMF)}, where the columns of $\bS$ are assumed to lie in the unit simplex.
SSMF has been found to be elegant and powerful---as shown by more than a decade of research on
{\em hyperspectral unmixing} (HU) in geoscience and remote sensing \cite{Jose12,Ken14SPM_HU}, and more recently, 
by
research in areas such as computer vision, machine learning, text mining and optimization~\cite{gillis2014and}.

To describe SSMF and its underlying significance, it is necessary to mention two key research topics from which important SSMF techniques were developed.
The first is HU, a main research topic in hyperspectral remote sensing.
The task of HU is to decompose a remotely sensed hyperspectral image into endmember spectral signatures and the corresponding abundance maps,
and SSMF plays the role of tackling such a decomposition.
A widely accepted assumption in HU is that $\bS$ has columns lying in the unit simplex;
or, some data pre-processing may be applied to make the aforementioned assumption happen \cite{chan2008convex,ma2010convex,Jose12,gillis2014and}.
Among the many SSMF techniques established within the  hyperspectral remote sensing community,
we should mention pure-pixel search
and minimum volume enclosing simplex (MVES) \cite{boardman1995mapping,nascimento2005vertex,Craig1994,Chan2009,Li2008,Lin2015icasspHyperCSI}---they are insightful and are recently shown to be theoretically sound \cite{Chan2011,gillis2014fast,lin2015identifiability}.

The second topic that SSMF has shown impact is topic discovery for text mining---which has recently received much interest in machine learning.
In this context,
the so-called separable NMF techniques have attracted considerable attention \cite{arora2012computing,arora2013practical,recht2012factoring,gillis2013robustness,fu2016robustness,fu2015self,esser2012convex,elhamifar2012see}.
Separable NMF falls into the scope of SSMF as it also assumes that the columns of $\bS$ lie in the unit simplex.
Separable NMF is very closely related to, if not exactly the same as, pure-pixel search developed earlier in HU; the two use essentially the same model assumption.
However, separable NMF offers new twists not seen in traditional HU, such as convex optimization solutions and {robustness analysis in the noisy case}; see the aforementioned references for details.
{Some recent research also considers more relaxed techniques than separable NMF, such as subset-separable NMF \cite{ge2015intersecting} and MVES \cite{huang2016anchor}.}
Furthermore, it is worth noting that 
other than HU and topic discovery, SSMF also find applications in various areas such as
gene expression data analysis,
dynamic biomedical imaging, and analytical chemistry \cite{wang2016mathematical,chen2011tissue,Lopes2010}.

The beauty of the aforementioned SSMF frameworks lies in how they utilize the geometric structures of the SSMF model to pin down sufficient conditions for exact recovery, 
and to build algorithms with good recovery performance.
We will shed some light onto those geometric insights when we review the problem in the next section,
and
we should note that
recent theoretical
breakthroughs in SSMF
have played a key role
in
understanding the fundamental natures of SSMF better and in designing better algorithms.
Motivated by such exciting advances, in this paper we explore a new theoretical direction for SSMF.
Our idea is still geometrical, but we use a different way, namely, by considering the maximum volume ellipsoid inscribed in a data-constructed convex hull;
the intuition will be elucidated later.
As the main contribution of this paper,
we will show a sufficient condition under which this maximum volume inscribed ellipsoid (MVIE) framework achieves exact recovery.
The sufficient recovery condition we prove is arguably not hard to satisfy in practice and is much 
{more relaxed}
than that of pure-pixel search and separable NMF, and {coincidentally} it is the same as that of MVES---which is a powerful SSMF framework for non-separable problem instances.
In addition, our development will reveal that MVIE can be practically realized by solving a facet enumeration problem, and then by solving a convex optimization problem in form of log determinant maximization.
This shows a very different flavor from the MVES framework in which we are required to solve a non-convex problem.
While we should point out that our MVIE solution
{may not be}
computed in polynomial time because 
{facet enumeration is NP-hard in general \cite{barot2017concise,boros2009generating},}
it still brings a new perspective to the SSMF problem.
In particular, for instances where facet enumeration can be efficiently computed, the remaining problem with MVIE is to solve a convex problem in which local minima are no longer an issue.
We will provide numerical results to 
{show}
the potential of the MVIE framework.

The organization of this paper is as follows.
We succinctly review the SSMF model and some existing frameworks in Section \ref{sec:modelandwork}.
The MVIE framework is described in Section \ref{sec:MVIE}.
Section \ref{sec:proof} provides the proof of the main theoretical result in this paper.
Section \ref{sec:algorithm} develops an MVIE algorithm and discusses computational issues.
Numerical results are provided in Section \ref{sec:simulation},
and we conclude this work in Section \ref{sec:conclusion}.


Our notations are standard, and some of them are specified as follows.
Boldface lowercase and capital letters, like $\ba$ and $\bA$, represent vectors and matrices, respectively (resp.);
unless specified, $\ba_i$ denotes the $i$th column of $\bA$;
$\be_i$ denotes a unit vector with $[\be_i]_i = 1$ and $[ \be_i ]_j = 0$ for $j \neq i$;
$\bone$ denotes an all-one vector;
$\ba \geq \bzero$ means that $\ba$ is element-wise non-negative;
the pseudo-inverse of a given matrix $\bA$ is denoted by $\bA^\dag$;
$\| \cdot \|$ denotes the Euclidean norm (for both vectors and matrices);
given a set $\setC$ in $\Rbb^n$, $\aff \ \setC$ and $\conv \ \setC$ denote the affine hull and convex hull of $\setC$, resp.;
the dimension of a set $\setC$ is denoted by $\dim \setC$;
$\inte \ \setC, \ri \ \setC, \bdy \ \setC$ and $\rbd \ \setC$ denote the interior, relative interior, boundary and relative boundary of the given set $\setC$, resp.;
$\vol \ \setC$ denotes the volume of a measurable set $\setC$;
$\setB_n = \{ \bx \in \Rbb^n ~|~ \| \bx \| \leq 1 \}$ denotes the $n$-dimensional unit Euclidean-norm ball, or simply unit ball;
$\Sbb^n$ and $\Sbb^n_+$ denote the sets of all $n \times n$ symmetric and symmetric positive semidefinite matrices, resp.;
$\lambda_{\rm min}(\bX)$ and $\lambda_{\rm max}(\bX)$ denote the smallest and largest eigenvalues of $\bX$, resp.

\section{Data Model and Related Work}\label{sec:modelandwork}

In this section we describe the background of SSMF.

\subsection{Model}

As mentioned in the Introduction, we consider a low-rank data model $$\bX = \bA \bS,$$
where $\bA \in \Rbb^{M \times N}, \bS \in \Rbb^{N \times L}$ with $N  \leq \min\{M,L\}$.
The model can be written in a column-by-column form as
\[
\bx_i = \bA \bs_i, \quad i=1,\ldots,L,
\]
and we assume that
\begin{itemize}
\item[(A1)] every $\bs_i$ lies in the unit simplex, i.e., $\bs_i \geq \bzero, \bone^T \bs_i = 1$;
\item[(A2)] $\bA$ has full column rank;
\item[(A3)] $\bS = [~ \bs_1, \ldots, \bs_L ~]$ has full row rank.
\end{itemize}
{The above assumptions} will be assumed without explicit mentioning in the sequel.
The problem is to recover $\bA$ and $\bS$ from the data points $\bx_1,\ldots,\bx_L$.
Since $\bs_i$'s lie in the unit simplex, we call this problem simplex-structured matrix factorization, or SSMF in short.
We will focus only on the recovery of $\bA$;
{once $\bA$ is retrieved,
the factor $\bS$ can simply be recovered by solving the inverse problems
\[
\min_{\bs_i \geq \bzero, \bone^T \bs_i = 1} ~ \| \bx_i - \bA \bs_i \|^2,
\qquad i=1,\ldots,L.
\]
}
SSMF finds many important applications as we reviewed in the Introduction,
and one can find an enormous amount of literature---from remote sensing, signal processing, machine learning, computer vision, optimization, etc.---on the wide variety of techniques for SSMF or related problems.
Here we selectively and concisely describe two mainstream frameworks.

\subsection{Pure-Pixel Search and Separable NMF}

\begin{figure}[t]
	\centerline{\resizebox{0.8\textwidth}{!}{\includegraphics{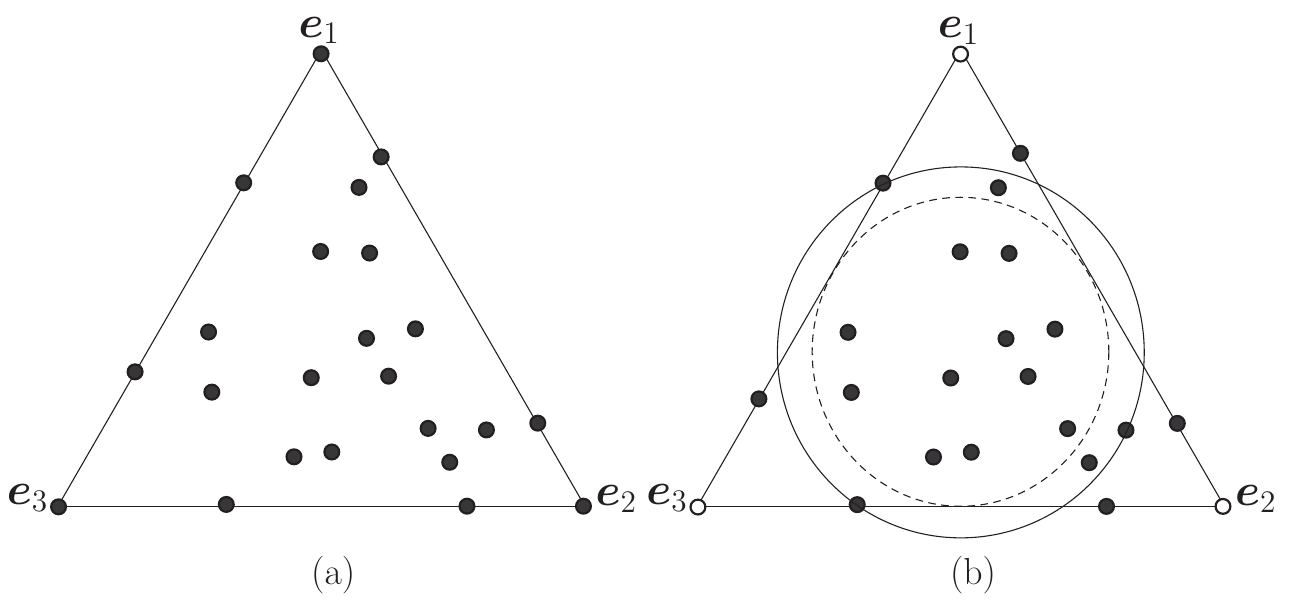}}}
    \caption{Illustration of geometry of the points $\bs_1,\ldots,\bs_L$.
    As an example we consider $N=3$,
    and {we view $\bs_1,\ldots,\bs_L$ by projecting them onto the ($2$-dimensional) affine hull $\aff\{ \be_1,\be_2,\be_3 \}$}.
    The solid dark dots are the $\bs_i$'s,
    and the solid line outlines the relative boundary of the unit simplex $\conv\{ \be_1,\be_2,\be_3 \}$.
    (a) The pure-pixel or separable case, in which $\be_1,\be_2,\be_3$ exist in some of $\bs_i$'s.
    (b) A no-pure-pixel or non-separable case in which $\gamma > 1/\sqrt{N-1}$ holds.
    {The dashed and solid circles correspond to the balls $(1/\sqrt{N-1}) \setB_N$ and $\gamma\setB_N$, resp.}
    }
    \label{fig:identifiability}
\end{figure}

The first framework to be reviewed is 
pure-pixel search in HU in remote sensing \cite{Ken14SPM_HU} or separable NMF in machine learning \cite{gillis2014and}.
Both assume that for every $k \in \{1,\ldots,N \}$, there exists an index $i_k \in \{1,\ldots,N \}$ such that
\[
\bs_{i_k} = \be_k.
\]
The above assumption is called the pure-pixel assumption in HU or separability assumption in separable NMF.
Figure \ref{fig:identifiability}(a) illustrates the geometry of $\bs_1,\ldots,\bs_L$ under the pure-pixel assumption,
where we see that the pure pixels $\bs_{i_1},\ldots,\bs_{i_N}$ are the vertices of the convex hull $\conv\{ \bs_1,\ldots, \bs_L \}$.
This suggests that some kind of vertex search can lead to recovery of $\bA$---the key insight of almost all algorithms in this framework.
The beauty of pure-pixel search or separable NMF is that under the pure-pixel assumption,
SSMF can be accomplished either via simple algorithms \cite{arora2013practical,fu2015self} or via convex optimization \cite{recht2012factoring,gillis2013robustness,fu2016robustness,esser2012convex,elhamifar2012see}.
Also, as shown in the aforementioned references, some of these algorithms are supported by theoretical analyses in terms of guarantees on recovery accuracies.

\begin{figure}[htp!]
	\begin{center}
		\resizebox{0.8\textwidth}{!}{\includegraphics{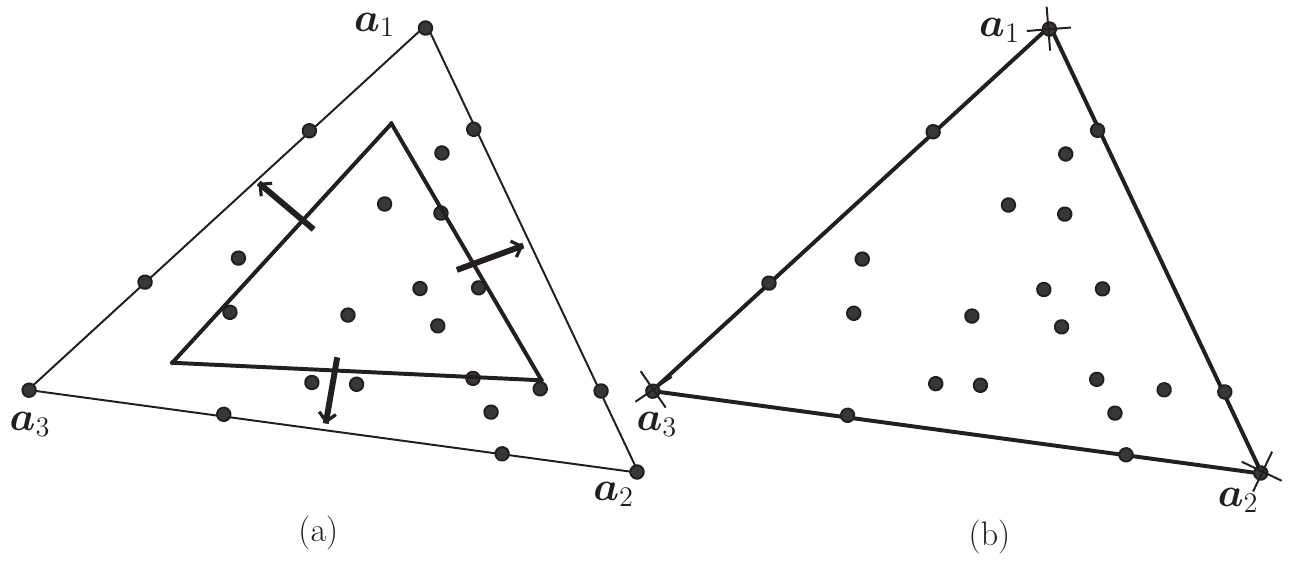}}
	\end{center}
	\caption{{Geometrical illustration of MVIS.
	The instance shown satisfies the pure-pixel assumption.
	The way we visualize is similar to that in Figure~\ref{fig:identifiability},
	where we project the data points $\bx_1,\ldots,\bx_L$ onto the affine hull $\aff\{\ba_1,\ba_2,\ba_3\}$.
	The solid dark dots are the data points $\bx_1,\ldots,\bx_L$.
	The subfigure in (a) depicts a simplex inscribed in the data convex hull $\conv\{ \bx_1,\ldots,\bx_L \}$.
	The outer triangle represents $\conv\{ \bx_1,\ldots,\bx_L \}$, while the inner triangle the inscribed simplex.
	The subfigure in (b) depicts the MVIS. 
	The vertices of the MVIS, marked by ``$\times$'', are seen to be $\ba_1,\ba_2,\ba_3$.
}}
	\label{fig:MVIS}
\end{figure}

{
	To give insights into how the geometry of the pure-pixel case can be utilized for SSMF,
	we briefly describe a pure-pixel search framework based on {\em maximum volume inscribed simplex} (MVIS) \cite{nascimento2005vertex,Chan2009}.
	The MVIS framework considers the following problem
	\begin{equation} \label{eq:mvis}
	\begin{aligned}
	\max_{\bb_1,\ldots,\bb_N \in \Rbb^M} & ~ \vol( \conv\{\bb_1,\ldots,\bb_N\} ) \\
	{\rm s.t.} & ~ \conv\{\bb_1,\ldots,\bb_N\} \subseteq \conv\{ \bx_1,\ldots, \bx_L \},
	\end{aligned}
	\end{equation}
	where we seek to find a simplex $\conv\{\bb_1,\ldots,\bb_N\}$ 
	such that it is inscribed in the data convex hull $\conv\{ \bx_1,\ldots, \bx_L \}$ and its volume is the maximum; see Figure~\ref{fig:MVIS} for an illustration.
	Intuitively, it seems true that the vertices of the MVIS,  under the pure-pixel assumption, should be $\ba_1,\ldots,\ba_N$.
	In fact, this can be shown to be valid:
	\begin{Theorem}
		\cite{Chan2009}
		\label{thm:mvis}
		The optimal solution to the MVIS problem \eqref{eq:mvis} is $\ba_1,\ldots,\ba_N$ or their permutations if and only if the pure-pixel assumption holds.
	\end{Theorem}
	It should be noted that the above theorem also reveals that the MVIS cannot correctly recover $\ba_1,\ldots,\ba_N$ for no-pure-pixel or non-separable problem instances.
	Readers are also referred to \cite{Chan2009} for details on how the MVIS problem is handled in practice.
}

\subsection{Minimum Volume Enclosing Simplex}

\begin{figure}[h!]
    \centerline{\resizebox{0.5\textwidth}{!}{\includegraphics{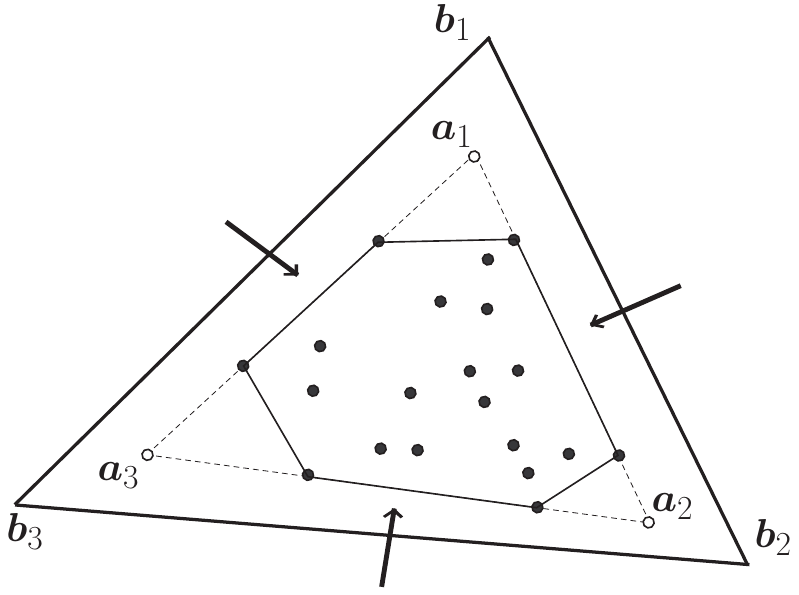}}}
    \caption{Geometrical illustration of MVES.
    The instance shown does not satisfy the pure-pixel assumption.
    The way we visualize is the same as that in Figure~\ref{fig:MVIS}.
    The solid dark dots are the data points $\bx_1,\ldots,\bx_L$,
    the dashed line outlines where is $\conv\{ \ba_1,\ba_2,\ba_3 \}$,
    the solid line inside $\conv\{ \ba_1,\ba_2,\ba_3 \}$ shows the relative boundary of the data convex hull $\conv\{ \bx_1,\ldots,\bx_L \}$,
    and the solid line outside $\conv\{ \ba_1,\ba_2,\ba_3 \}$ shows the relative boundary of  a data-enclosing simplex $\conv\{ \bb_1,\bb_2,\bb_3 \}$.
    From this illustration it seems likely that the minimum volume data-enclosing simplex would be $\conv\{ \ba_1,\ba_2,\ba_3 \}$ itself.
    }
    \label{fig:MVES}
\end{figure}

While SSMF under the pure-pixel assumption gives many benefits,
the assumption of having pure pixels in the data is somewhat strong.
A question that has previously puzzled researchers
is whether recovery of $\bA$ is possible without the pure-pixel assumption.
This leads to another framework that hinges on minimum volume enclosing simplex (MVES)---a notion conceived first by Craig in the HU context \cite{Craig1994} and an idea that can be traced back to the 1980's \cite{Full81}.
The idea is to solve an MVES problem
\begin{equation} \label{eq:mves}
\begin{aligned}
\min_{\bb_1,\ldots,\bb_N \in \Rbb^M} & ~ \vol( \conv\{\bb_1,\ldots,\bb_N\} ) \\
{\rm s.t.} & ~ \bx_i \in \conv\{\bb_1,\ldots,\bb_N\}, \quad i=1,\ldots,L,
\end{aligned}
\end{equation}
or its variants (see, e.g., \cite{Dias2009,fu2016robust}).
As can be seen in \eqref{eq:mves} and as illustrated in Figure \ref{fig:MVES},
the goal is to find a simplex that encloses the data points and has the minimum volume.
The vertices of the MVES, which is the solution $\bb_1,\ldots,\bb_N$ to Problem \eqref{eq:mves}, then serves as the estimate of $\bA$.
MVES is more commonly seen in HU, and most recently the idea has made its way to machine learning \cite{huang2016anchor,fu2015blind}.
Empirically it has been observed that MVES can achieve good recovery accuracies in the absence of pure pixels,
and MVES-based algorithms are often regarded as tools for resolving instances of ``heavily mixed pixels'' in HU \cite{miao2007endmember}.
Recently, the mystery of whether MVES can provide exact recovery {\em theoretically} has been answered:

\begin{Theorem}
\cite{lin2015identifiability}
\label{thm:mves}
Define
\begin{equation} \label{eq:uppl}
\gamma= \max \left\{ r \leq  1 ~|~ ( \conv\{ \be_1,\ldots,\be_N \} ) \cap ( r \setB_N  ) \subseteq \conv\{\bs_1,\ldots,\bs_L\} \right\},
\end{equation}
which is called the uniform pixel purity level.
If $N \geq 3$ and
$$\gamma > \frac{1}{\sqrt{N-1}},$$ then the optimal solution to the MVES problem \eqref{eq:mves} must be given by $\ba_1,\ldots,\ba_N$ or their permutations.
\end{Theorem}
The uniform pixel purity level has elegant geometric interpretations.
To give readers some feeling,
Figure \ref{fig:identifiability}(b) illustrates an instance for which $\gamma > 1/\sqrt{N-1}$ holds, but the pure-pixel assumption does not.
Also, note that $\gamma=1$ corresponds to the pure-pixel case.
Interested readers are referred to \cite{lin2015identifiability} for more explanations of $\gamma$, and \cite{huang2016anchor,fu2015blind,fu2016robust} for concurrent and more recent results for theoretical MVES recovery.
Loosely speaking,
the premise in Theorem~\ref{thm:mves} should have a high probability to satisfy in practice as far as the data points are reasonably well spread.

While MVES is appealing in its recovery guarantees,
the pursuit of SSMF frameworks is arguably not over.
The MVES problem \eqref{eq:mves} is non-convex and NP-hard in general \cite{packer2002np}.
Our numerical experience is that
the convergence of an MVES algorithm to a good result could depend on the starting point.
Hence, 
it is interesting to study alternative frameworks that can also go beyond  the pure-pixel or separability case and can bring new perspective to the no-pure-pixel case---and this is the motivation for our development of the MVIE framework in the next section.

\section{Maximum Volume Inscribed Ellipsoid}\label{sec:MVIE}

Let us first describe some facts and our notations with ellipsoids.
Any $n$-dimensional ellipsoid $\setE$ in $\Rbb^m$ may be characterized as
\[
\setE= \setE(\bF, \bc ) \triangleq \{ \bF \balp + \bc ~|~ \| \balp \| \leq 1 \},
\]
for some full column-rank $\bF \in \Rbb^{m \times n}$ and $\bc \in \Rbb^m$.
The volume of an $n$-dimensional ellipsoid $\setE(\bF, \bc )$ is given by
\[
\vol(\setE(\bF, \bc )) = \rho_n ( \det( \bF^T \bF) )^{1/2},
\]
where $\rho_n$ denotes the volume of the $n$-dimensional unit ball \cite{boyd2004convex}.

\begin{figure}[htp!]
    \centerline{\resizebox{0.8\textwidth}{!}{\includegraphics{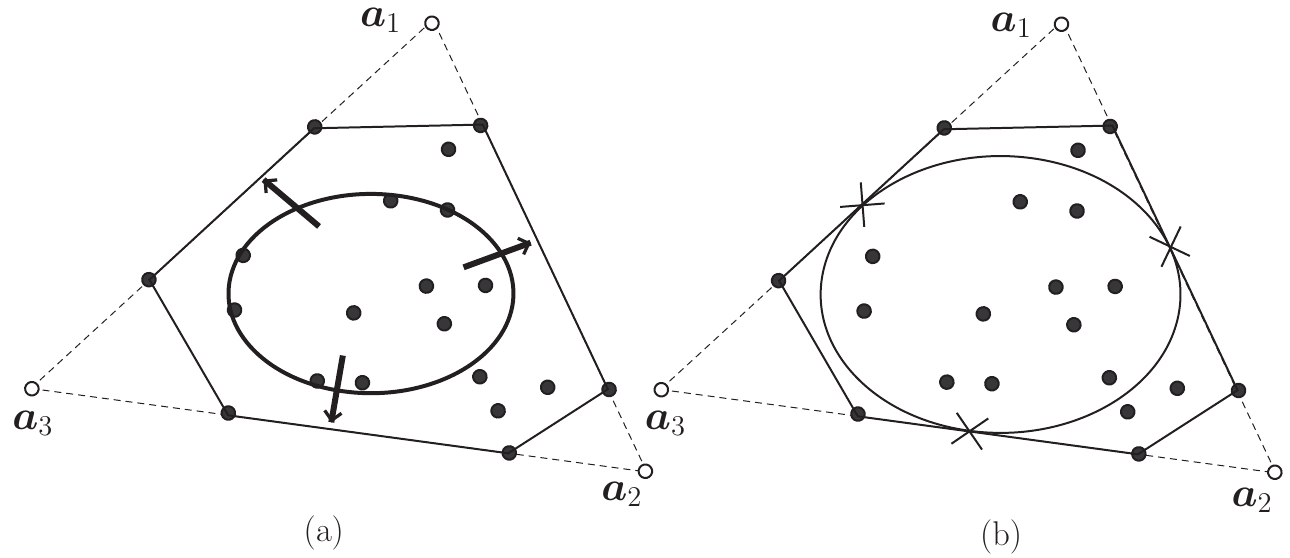}}}
    \caption{Geometrical illustration of MVIE.
    	The instance shown does not satisfy the pure-pixel assumption.
    	The way we visualize is the same as that in Figure~\ref{fig:MVIS}.
    In the subfigure (a), the circle depicts an ellipsoid inscribed in the data convex hull $\conv\{ \bx_1,\ldots,\bx_L \}$.
    The subfigure in (b) shows a possible scenario for which the MVIE has contact points with $\conv\{ \bx_1,\ldots,\bx_L \}$;
    those contact points are marked by ``$\times$''.
    }
    \label{fig:MVIE}
\end{figure}

We are interested in an MVIE problem whose aim is to find a maximum volume ellipsoid contained in the convex hull of the data points.
For convenience,
denote
\[
\setX = \conv\{ \bx_1,\ldots, \bx_L \}
\]
to be the convex hull of the data points.
As a basic result one can show that
\begin{equation} \label{eq:dimX}
\dim \setX  = \dim( \aff\{ \bx_1,\ldots,\bx_L \} ) =  \dim ( \aff\{ \ba_1,\ldots, \ba_N \} )=  N-1;
\end{equation}
note that the second equality is due to $\aff\{ \bx_1,\ldots,\bx_L \}= \aff\{ \ba_1,\ldots, \ba_N \}$ under (A3), which was proved in \cite{chan2008convex,Chan2009}.
Hence we also restrict the dimension of the ellipsoid to be $N-1$,
and the MVIE problem is formulated as
\begin{equation} \label{eq:mainP}
\begin{aligned}
\max_{\bF, \bc} ~ & \det( \bF^T \bF ) \\
{\rm s.t.} ~ &
\setE(\bF, \bc ) \subseteq \setX,
\end{aligned}
\end{equation}
where $\bF \in \Rbb^{M \times (N-1)}, \bc \in \Rbb^M$.\footnote{Notice that we do not constrain $\bF$ to be of full column rank in Problem \eqref{eq:mainP} for the following reasons. First, it can be verified that a feasible $\setE(\bF,\bc)$ with $\bF$ being of full column rank always exists if $\dim \setX = N-1$. Second, if $\bF$ does not have full column rank then $\det(\bF^T \bF)= 0$.}
{It is interesting to note that the MVIE formulation above is similar to the MVIS formulation \eqref{eq:mvis}; the inscribed simplex in MVIS is replaced by an ellipsoid.
However, the pursuit of MVIE leads to significant differences from that of MVIS.}
To see it, consider the illustration in Figure \ref{fig:MVIE}.
We observe that the MVIE and the data convex hull $\setX$ have contact points
{on their relative boundaries}.
{Since those contact points are also on the ``appropriate'' facets of $\conv\{ \ba_1, \ldots, \ba_N \}$ (for the instance in Figure \ref{fig:MVIE}), they may provide clues on how to recover $\ba_1, \ldots, \ba_N$.}

The following theorem describes the main result of this paper.

\begin{Theorem} \label{thm:main}
Suppose that $N \geq 3$ and $\gamma > 1/\sqrt{N-1}$.
The MVIE, or the optimal ellipsoid of Problem \eqref{eq:mainP}, is uniquely given by
\begin{equation} \label{eq:main_mvie_sol}
\setE^\star = \setE\left( \tfrac{1}{\sqrt{N(N-1)}} \bA \bC, \bar{\ba} \right),
\end{equation}
where $\bC \in \Rbb^{N \times (N-1)}$ is any semi-unitary matrix such that $\bC^T \bone = \bzero$, and $\bar{\ba} = \frac{1}{N} \sum_{i=1}^N \ba_i$.
Also, there are exactly $N$ contact points between $\setE^\star$ and $\rbd \ \setX$, that is,
\begin{equation} \label{eq:main_contact}
\setE^\star \cap ( \rbd \ \setX ) = \{ \bq_1, \ldots, \bq_N \},
\end{equation}
and those contact points are given by
\begin{equation} \label{eq:main_contact_sol}
\bq_i = \frac{1}{N-1} \sum_{j \neq i} \ba_j.
\end{equation}
\end{Theorem}
%

Theorem~\ref{thm:main} gives a vital implication on a condition under which we can leverage MVIE to exactly recover $\bA$. Consider the following corollary as a direct consequence of Theorem~\ref{thm:main}.

\begin{Corollary}\label{MainCorollary}
Under the premises of $N \geq 3$ and $\gamma > 1/\sqrt{N-1}$,
we can exactly recover $\bA$ by solving the MVIE problem \eqref{eq:mainP}, finding the contact points $\bq_i$'s in \eqref{eq:main_contact}, and reconstructing $\ba_i$'s either via
\[
\ba_i = N \bar{\ba} - (N-1) \bq_i, \quad i=1,\ldots,N,
\]
or via
\[
\ba_i = \sum_{j=1}^N \bq_j - (N-1) \bq_i, \quad i=1,\ldots,N.
\]
\end{Corollary}
Hence, we have shown a new and provably correct SSMF framework via MVIE.
Coincidentally and beautifully, the sufficient exact recovery condition of this MVIE framework is the same as that of the MVES framework (cf. Theorem~\ref{thm:mves})---which suggests that MVIE should be as powerful as MVES.

In the next section we will describe the proof of Theorem~\ref{thm:main}.
{We will also 
develop an algorithm for implementing MVIE,
}
and then testing it through numerical experiments; these will be considered in
Sections \ref{sec:algorithm}--\ref{sec:simulation}.

\section{Proof of Theorem~\ref{thm:main}}\label{sec:proof}

Before we give the full proof of Theorem~\ref{thm:main}, we should briefly mention the insight behind.
At the heart of our proof is John's theorem for MVIE characterization, which is described as follows.
\begin{Theorem}
\cite{gruber2005arithmetic} \label{thm:john} Let $\setT \subset \Rbb^n$ be a compact convex set with non-empty interior.
The following two statements are equivalent.
\begin{enumerate}[\leftmargin=1cm]
\item (a) The $n$-dimensional ellipsoid of maximum volume contained in $\setT$ is uniquely given by $\setB_n$.

\smallskip

\item (b) $\setB_n \subseteq \setT$ and there exist points $\bu_1,\ldots,\bu_r \in \setB_n \cap (\bdy \ \setT)$, with $r \geq n+1$, such that
\[
\sum_{i=1}^r \lambda_i \bu_i = \bzero, \qquad \sum_{i=1}^r \lambda_i \bu_i \bu_i^T = \bI,
\]
for some $\lambda_1,\ldots,\lambda_r > 0$.
\end{enumerate}
\end{Theorem}
There are however challenges to be overcome.
First, John's theorem cannot be directly applied to our MVIE problem \eqref{eq:mainP} because $\setX$ does not have an interior (although $\setX$ has non-empty relative interior).
Second, John's theorem does not tell us how to identify the contact points $\bu_i$'s---which we will have to find out.
Third, our result in Theorem~\ref{thm:main} is stronger in the sense that we characterize the set of {\em all} the contact points, and this will require some extra work.

The proof of Theorem~\ref{thm:main} is divided into three parts and described in the following subsections.
Before we proceed, let us define some specific notations that will be used throughout the proof.
We will denote an affine set by
\[
\setA(\bPhi,\bb) \triangleq \{ \bPhi \balp + \bb ~|~ \balp \in \Rbb^n \},
\]
for some $\bPhi \in \Rbb^{m \times n}, \bb \in \Rbb^n$.
In fact, any affine set $\setA$ in $\Rbb^m$ of $\dim \setA = n$ may be represented by $\setA = \setA(\bPhi,\bb)$ for some full column rank $\bPhi \in \Rbb^{m \times n}$ and $\bb \in \Rbb^m$.
Also, we let $\bC \in \Rbb^{N \times (N-1)}$ denote any matrix such that
\begin{equation} \label{eq:C_def}
\bC^T \bC = \bI, \quad \bC^T \bone = \bzero,
\end{equation}
and we let
\begin{equation} \label{eq:d_def}
\bd = \tfrac{1}{N} \bone \in \Rbb^N.
\end{equation}

\subsection{Dimensionality Reduction}

Our first task is to establish an equivalent MVIE transformation result.

\begin{Prop} \label{prop:dr}
Represent the affine hull $\aff\{ \bx_1,\ldots, \bx_L \}$ by
\begin{equation} \label{eq:prop1:affX}
\aff\{ \bx_1,\ldots, \bx_L \} = \setA(\bPhi,\bb)
\end{equation}
for some full column rank $\bPhi \in \Rbb^{M \times (N-1)}$ and $\bb \in \Rbb^M$.
Let
\[
\bx'_i = \bPhi^\dag ( \bx_i - \bb ), ~ i=1,\ldots,L, \qquad
\setX' = \conv\{ \bx_1', \ldots, \bx_L' \} \subset \Rbb^{N-1}.
\]
The MVIE problem \eqref{eq:mainP} is equivalent to
\begin{equation} \label{eq:drP}
\begin{aligned}
\max_{\bF', \bc'} ~ & |\det(\bF')|^2 \\
{\rm s.t.} ~ &
\setE(\bF', \bc' ) \subseteq \setX',
\end{aligned}
\end{equation}
where $\bF' \in \Rbb^{(N-1) \times (N-1)}$, $\bc' \in \Rbb^{N-1}$.
In particular, the following properties hold:
\begin{enumerate}[\leftmargin=1cm]
\item (a) If $(\bF, \bc)$ is a feasible (resp., optimal) solution to Problem \eqref{eq:mainP}, then
\begin{equation} \label{eq:prop1:sol1}
(\bF',\bc') = ( \bPhi^\dag \bF, \bPhi^\dag( \bc - \bb ) )
\end{equation}
is a feasible (resp., optimal) solution to Problem \eqref{eq:drP}.

\smallskip

\item (b) If $(\bF,' \bc')$ is a feasible (resp., optimal) solution to Problem \eqref{eq:drP}, then
\begin{equation} \label{eq:prop1:sol2}
(\bF,\bc) = ( \bPhi \bF', \bPhi \bc' + \bb)
\end{equation} is a feasible (resp., optimal) solution to Problem \eqref{eq:mainP}.

\smallskip

\item (c) The set $\setX'$ has non-empty interior.

\smallskip

\item (d) Let $(\bF,\bc)$  be a feasible solution to Problem~\eqref{eq:mainP},
and let $(\bF',\bc')$ be given by  \eqref{eq:prop1:sol1};
or, let $(\bF',\bc')$ be a feasible solution to Problem \eqref{eq:drP}, and let $(\bF, \bc)$ be given by \eqref{eq:prop1:sol2}.
Denote $\setE = \setE(\bF,\bc)$ and $\setE' = \setE(\bF',\bc')$.
Then
\begin{align*}
\bq \in \setE \cap ( \rbd \ \setX ) \quad & \Longrightarrow \quad \bq' = \bPhi^\dag ( \bq - \bb ) \in \setE' \cap (\bdy \ \setX' ), \\
\bq' \in \setE' \cap (\bdy \ \setX' ) \quad & \Longrightarrow \quad \bq = \bPhi\bq' + \bb \in \setE \cap ( \rbd \ \setX ).
\end{align*}

\end{enumerate}
\end{Prop}

The above result is a dimensionality reduction (DR) result where we equivalently transform the MVIE problem from a higher dimension space (specifically, $\Rbb^M$) to a lower dimensional space (specifically, $\Rbb^{N-1}$).
It has the same flavor as the so-called affine set fitting result in \cite{chan2008convex,Chan2009}, which is also identical to principal component analysis.
This DR result will be used again when we develop an algorithm for MVIE in later sections.
We relegate the proof of Proposition~\ref{prop:dr} to Appendix \ref{sec:proof-prop1}.

Now, we construct an equivalent MVIE problem via a specific choice of $(\bPhi,\bb)$.
It has been shown that under (A3),
\begin{equation} \label{eq:proof2_t1}
\aff\{ \bx_1,\ldots, \bx_L \} = \aff\{ \ba_1,\ldots,\ba_N \};
\end{equation}
see \cite{chan2008convex,Chan2009}.
Also, consider  the following fact.
\begin{Fact}\cite{lin2015identifiability} \label{fac:aff_unit}
The affine hull of all unit vectors $\be_1,\ldots, \be_N$ in $\Rbb^N$ can be characterized as
\[
\aff\{ \be_1,\ldots, \be_N \} = \setA(\bC,\bd),
\]
{where $\bC$ and $\bd$ have been defined in \eqref{eq:C_def} and \eqref{eq:d_def}, resp.}
\end{Fact}
Applying Fact~\ref{fac:aff_unit} to \eqref{eq:proof2_t1} yields
\[
\aff\{ \bx_1,\ldots, \bx_L \} = \setA(\bA \bC, \bA \bd ).
\]
By choosing $(\bPhi,\bb) = (\bA\bC, \bA \bd)$ and applying Proposition \ref{prop:dr},
we obtain an equivalent MVIE problem in \eqref{eq:drP} that has
\[
\bx_i = \bA \bC \bx_i' + \bA \bd, \quad i=1,\ldots,L.
\]
The above equation can be simplified.
By plugging the model $\bx_i = \bA \bs_i$ into the above equation, we get
$\bs_i = \bC \bx_i' + \bd$;
and using the properties $\bC^T \bC = \bI$ and $\bC^T \bd = \bzero$ we further get
$\bx_i' = \bC^T \bs_i.$
By changing the notation $\setX'$ to $\setS'$, and $\bx_i'$ to $\bs_i'$, we rewrite the equivalent MVIE problem \eqref{eq:drP} as
\begin{equation} \label{eq:drP2}
\begin{aligned}
\max_{\bF', \bc'} ~ & | \det(\bF') |^2 \\
{\rm s.t.} ~ &
\setE(\bF', \bc' ) \subseteq \setS',
\end{aligned}
\end{equation}
where we again have $\bF' \in \Rbb^{(N-1) \times (N-1)}$, $\bc' \in \Rbb^{N-1}$;
$\setS'$ is given by
$\setS' = \conv\{ \bs_1',\ldots,\bs_L' \}$ with
\[ \bs_i' = \bC^T \bs_i, \quad i=1,\ldots,L. \]
Furthermore, note that $\setS'$ has non-empty interior; cf. Statement (c) of Proposition \ref{prop:dr}.

\subsection{Solving the MVIE via John's Theorem}

Next, we apply John's theorem to the equivalent MVIE problem in \eqref{eq:drP2}.
{It would be helpful to first describe the outline of our proof.}
For convenience, let
$$\beta = \frac{1}{\sqrt{N(N-1)}}$$
and
\[
\bq_i' = \frac{1}{N-1} \sum_{j \neq i} \bC^T \be_j, \quad i=1,\ldots,N.
\]
We will show that the optimal ellipsoid to Problem  \eqref{eq:drP2} is uniquely given by $\beta \setB_{N-1}$,
and that $\bq_1', \ldots, \bq_N'$ lie in $(\beta \setB_{N-1}) \cap (\bdy \ \setS')$;
the underlying premise is $\gamma \geq 1/\sqrt{N-1}$.
Subsequently, by the equivalence properties in Proposition \ref{prop:dr}, and by $\beta \setB_{N-1} = \setE( \beta \bI, \bzero)$, we have
\begin{equation}\label{eq:def:Estar}
\setE( \beta \bA \bC, \bA \bd ) = \setE^\star
\end{equation}
as the optimal ellipsoid of our original MVIE problem \eqref{eq:mainP}; also, we have
\[
\bq_i  = \bA \bC \bq'_i + \bA \bd \in \setE^\star \cap ( \rbd \ \setX ), \quad i=1,\ldots,N.
\]
Furthermore, it will be shown that $\bq_i$ can be reduced to $\bq_i = \tfrac{1}{N-1} \sum_{j \neq i} \ba_j$.
Hence, except for the claim $\{ \bq_1,\ldots,\bq_N \} = \setE^\star \cap ( \rbd \ \setX )$, we see all the results in Theorem~\ref{thm:main}.

Now, we show the more detailed parts of the proof.

\medskip
{\em Step 1:} \
Let us assume $\beta \setB_{N-1} \subseteq \setS'$ and $\bq'_i \in (\beta \setB_{N-1}) \cap ( \bdy \ \setS' )$ for all $i$; we will come back to this later.
The aim here is to verify that $\beta \setB_{N-1}$ and $\bq_1',\ldots,\bq_N'$ satisfy the MVIE conditions in John's theorem.
Since $\bC^T \bone = \bzero$, we can simplify $\bq_i'$ to
\[
\bq_i' = \frac{1}{N-1} \bC^T ( \bone - \be_i ) = - \frac{1}{N-1} \bC^T \be_i.
\]
Consequently, one can verify that
\begin{align*}
(N-1)^2 \sum_{i=1}^N \bq_i' & = -{(N-1)} \bC^T \bone = \bzero, \\
(N-1)^2 \sum_{i=1}^N (\bq_i')(\bq'_i)^T & = \bC^T \left( \sum_{i=1}^N \be_i \be_i^T \right) \bC = \bC^T \bI \bC = \bI,
\end{align*}
which are the MVIE conditions of John's theorem; see Statement (b) of Theorem~\ref{thm:john},
{with $\bu_i = \bq_i'$, $\lambda_i = (N-1)^2$, $i=1,\ldots,N$.}
Hence, $\beta \setB_{N-1}$ is the unique maximum volume ellipsoid contained in $\setS'$.

\medskip
{\em Step 2:} \
We verify that $\beta \setB_{N-1} \subseteq \setS'$ if $\gamma \geq 1/\sqrt{N-1}$.
The verification requires another equivalent MVIE problem, given as follows:
\begin{equation} \label{eq:drP3}
\begin{aligned}
\max_{\bF, \bc} ~ & \det(\bF^T \bF) \\
{\rm s.t.} ~ &
\setE(\bF, \bc ) \subseteq \setS,
\end{aligned}
\end{equation}
where
\[ \setS = \conv\{ \bs_1,\ldots, \bs_L \}, \]
and with a slight abuse of notations we redefine  $\bF \in \Rbb^{N \times (N-1)}$, $\bc \in \Rbb^N$.
Using the same result in the previous subsection,
it can be readily shown that Problem \eqref{eq:drP3} is equivalent to Problem \eqref{eq:drP2} under $(\bPhi,\bb) = (\bC,\bd)$.
Let
\[
\setE = \setE \left( \beta \bC, \bd \right), \qquad \setE' =  \setE\left( \beta \bI, \bzero \right) = \beta \setB_{N-1}.
\]
From Statement (a) of Proposition \ref{prop:dr}, we have $\setE \subseteq \setS \Longrightarrow \setE' \subseteq \setS'$;
thus, we turn to proving $\setE \subseteq \setS$.
Recall from the definition of $\gamma$ in \eqref{eq:uppl} that
\begin{equation} \label{eq:pf_0_t1}
( \conv\{ \be_1,\ldots, \be_N \} ) \cap ( \gamma \setB_N ) \subseteq \setS.
\end{equation}
For $\gamma \geq 1/\sqrt{N-1}$, \eqref{eq:pf_0_t1} implies
\begin{equation} \label{eq:pf_0_t2}
( \conv\{ \be_1,\ldots, \be_N \} ) \cap \left( \tfrac{1}{\sqrt{N-1}} \setB_N \right) \subseteq \setS.
\end{equation}
Consider the following fact.
\begin{Fact}\cite{lin2015identifiability} \label{fac:dunno}
The following results hold.
\begin{enumerate}[\leftmargin=1cm]
\item (a) $( \aff\{ \be_1, \ldots, \be_N \} ) \cap \left( r \setB_N \right) = \setE\left( \sqrt{r^2-\tfrac{1}{N}} \bC, \bd \right)$ for $r \geq \frac{1}{\sqrt{N}}$;

\smallskip

\item (b) $( \conv\{ \be_1, \ldots, \be_N \} ) \cap \left( r \setB_N \right) = \aff\{ \be_1, \ldots, \be_N \} \cap \left( r \setB_N \right)$ for $\frac{1}{\sqrt{N}} < r \leq \frac{1}{\sqrt{N-1}}$.
\end{enumerate}
\end{Fact}
Applying Fact~\ref{fac:dunno} to the left-hand side of \eqref{eq:pf_0_t2} yields
\begin{equation} \label{eq:pf_0_t3}
( \conv\{ \be_1,\ldots, \be_N \} ) \cap \left( \frac{1}{\sqrt{N-1}} \setB_N \right) = \setE\left( \beta \bC, \bd \right).
\end{equation}
Hence, we have $\setE = \setE\left( \beta \bC, \bd \right) \subseteq \setS$,
which implies that $\beta \setB_{N-1}  = \setE' \subseteq \setS'$.

\medskip
{\em Step 3:} \
We verify that $\bq'_i \in (\beta \setB_{N-1}) \cap ( \bdy \ \setS' )$ for all $i$.
Again, the verification is based on the equivalence of Problem \eqref{eq:drP3} and Problem \eqref{eq:drP2} used in Step 2.
Let
\begin{equation}
\label{eq:proof:tt2}
\bw_i = \frac{1}{N-1} \sum_{j \neq i} \be_j, \quad i=1,\ldots,N,
\end{equation}
and let $\bw_i' = \bC^T (\bw_i - \bd)$ for all $i$.
By Statement (d) of Proposition \ref{prop:dr},
we have $\bw_i \in \setE \cap ( \rbd \ \setS ) \Longrightarrow \bw_i' \in \setE' \cap ( \bdy \ \setS' )$.
Also, owing to $\bC^T \bd = \bzero$, we see that $\bw_i' = \bC^T ( \tfrac{1}{N-1} \sum_{j\neq i} \be_j ) = \bq_i'$.
Hence, we can focus on showing $\bw_i \in \setE \cap ( \rbd \ \setS )$.
Since $\bw_i \in \aff\{ \be_1,\ldots, \be_N \} = \setA(\bC,\bd)$ (cf. Fact~\ref{fac:aff_unit}),
we can represent $\bw_i$ by
\begin{equation} \label{eq:proof:tt1}
\bw_i = \bC \bw_i' + \bd.
\end{equation}
Using \eqref{eq:proof:tt2}, $\bC^T \bC = \bI$ and $\bC^T \bd = \bzero$, one can verify that
\[
\frac{1}{N-1} = \| \bw_i \|^2
= \| \bC \bw_i' \|^2 + \| \bd \|^2 = \| \bw_i' \|^2 + \frac{1}{N},
\]
which is equivalent to $\| \bw_i' \| = \beta$.
We thus have $\bw_i \in \setE( \beta \bC, \bd) = \setE$.
Since $\setE \subseteq \setS$ (which is shown in Step 2), we also have $\bw_i \in \setS$.
The vector $\bw_i$ has $[ \bw_i ]_i = 0$, and as a result $\bw_i$ must not lie in $\ri \ \setS$.
It follows that $\bw_i \in \rbd \ \setS$.

\medskip
{\em Step 4:} \ Steps 1--3 essentially prove all the key components of the big picture proof described in the beginning of this subsection.
In this last step, we show the remaining result, namely, $\bq_i  = \bA \bC \bq'_i + \bA \bd = \tfrac{1}{N-1} \sum_{j \neq i} \ba_j$.
In Step 3, we see from $\bw_i' = \bq_i'$ and  \eqref{eq:proof:tt2}--\eqref{eq:proof:tt1} that $\bC \bq_i' + \bd = \tfrac{1}{N-1} \sum_{j \neq i} \be_j$.
Plugging this result into $\bq_i$ yields the desired result.

\subsection{On the Number of Contact Points}

Our final task is to prove that $\{ \bq_1,\ldots,\bq_N \} = \setE^\star \cap ( \rbd \ \setX )$;
note that the previous proof allows us only to say that  $\{ \bq_1,\ldots,\bq_N \} \subseteq \setE^\star \cap ( \rbd \ \setX )$.
We use the equivalent MVIE problem \eqref{eq:drP3} to help us solve the problem.
Again, let $\setE = \setE( \beta \bC, \bd )$ for convenience.
The crux is to show that
\begin{equation} \label{eq:pf3_key}
\bw \in \setE \cap (\rbd \ \setS ) 	   \quad \Longrightarrow \quad \bw = \bw_i ~ \text{for some $i \in \{1,\ldots,N\}$,}
\end{equation}
where $\bw_i$'s have been defined in \eqref{eq:proof:tt2}; the premise is $\gamma > 1/\sqrt{N-1}$.
By following the above development, especially, the equivalence results of Problems~\eqref{eq:drP3} and \eqref{eq:drP2}
and those of Problems~\eqref{eq:mainP} and \eqref{eq:drP2},
it can be verified that \eqref{eq:pf3_key} is equivalent to
\[
\bq \in \setE^\star \cap (\rbd \ \setX ) 	   \quad \Longrightarrow \quad \bq = \bq_i ~ \text{for some $i \in \{1,\ldots,N\}$,}
\]
which completes the proof of $\{ \bq_1,\ldots,\bq_N \} = \setE^\star \cap ( \rbd \ \setX )$.
We describe the proof of \eqref{eq:pf3_key} as follows.

\medskip
{\em Step 1:} \
First,
we show the following implication under $\gamma > 1/\sqrt{N-1}$:
\begin{equation} \label{eq:pf3_key1}
\bw \in \setE \cap (\rbd \ \setS ) 	   \quad \Longrightarrow \quad
\bw \in \setE \cap ( \rbd ( \conv\{ \be_1,\ldots,\be_N \} ) ).
\end{equation}
The proof is as follows. Let
\[
\setR(\gamma) = ( \conv\{ \be_1,\ldots,\be_N \} ) \cap ( \gamma \setB_N ),
\]
and note from \eqref{eq:pf_0_t1}--\eqref{eq:pf_0_t3} that
\begin{equation} \label{eq:pf3_1_t2}
\setE \subseteq \setR(\gamma) \subseteq \setS
\end{equation}
holds for $\gamma \geq 1/\sqrt{N-1}$.
It can be seen or easily verified from the previous development that
\begin{equation} \label{eq:pf3_1_t1}
\aff \ \setE = \aff \ \setS = \aff ( \conv\{ \be_1,\ldots, \be_N \} ) = \aff\{ \be_1,\ldots, \be_N \} = \setA(\bC,\bd).
\end{equation}
Also, by applying \eqref{eq:pf3_1_t1} to \eqref{eq:pf3_1_t2}, we get
$\aff( \setR(\gamma) ) = \setA(\bC,\bd)$.
It is then immediate that
\begin{equation} \label{eq:pf3_1_t3}
\ri( \setR(\gamma) ) \subseteq \ri \ \setS.
\end{equation}
From \eqref{eq:pf3_1_t2}--\eqref{eq:pf3_1_t3} we observe that
\begin{equation} \label{eq:pf3_1_t4}
\bw \in \setE, ~ \bw \in \rbd \ \setS
	\quad \Longrightarrow \quad
	\bw \in \setR(\gamma), ~ \bw \notin \ri(\setR(\gamma))  \quad \Longrightarrow \quad
	\bw \in \rbd(\setR(\gamma)).
\end{equation}
Let us further examine the right-hand side of the above equation.
For $\gamma > 1/\sqrt{N}$, we can write
\begin{align*}
\setR(\gamma) & = ( \conv\{ \be_1,\ldots,\be_N \} ) \cap \left( \aff\{ \be_1,\ldots,\be_N \} \cap (\gamma \setB_N)    \right)  \\
	& = ( \conv\{ \be_1,\ldots,\be_N \} ) \cap \left(  \setE\left(  \sqrt{\gamma^2 - \tfrac{1}{N}} \bC, \bd   \right)\right),
\end{align*}
where the second equality is due to Fact~\ref{fac:dunno}.(a).
It follows that
\begin{equation} \label{eq:pf3_1_t5}
\bw \in \rbd(\setR(\gamma))   \quad \Longrightarrow \quad
\bw \in \rbd(  \conv\{ \be_1,\ldots,\be_N \} ) \text{~or~} \bw \in \rbd \left(  \setE\left(  \sqrt{\gamma^2 - \tfrac{1}{N}} \bC, \bd   \right)\right).
\end{equation}
However, for $\gamma > 1/\sqrt{N-1}$, we have
\begin{equation} \label{eq:pf3_1_t6}
\bw \in \setE = \setE(\beta \bC, \bd)  = \setE\left(\sqrt{ \tfrac{1}{N-1} - \tfrac{1}{N} }  \bC, \bd \right)
\quad \Longrightarrow \quad
\bw \notin \rbd \left(  \setE\left(  \sqrt{\gamma^2 - \tfrac{1}{N}} \bC, \bd   \right)\right).
\end{equation}
By combining \eqref{eq:pf3_1_t4}, \eqref{eq:pf3_1_t5} and \eqref{eq:pf3_1_t6}, we obtain \eqref{eq:pf3_key1}.

\medskip
{\em Step 2:} \
Second, we show that
\begin{equation} \label{eq:pf3_key2}
\bw \in \setE \cap ( \rbd ( \conv\{ \be_1,\ldots,\be_N \} ))
	\quad \Longrightarrow \quad
	\bw = \bw_i ~ \text{for some $i \in \{1,\ldots,N\}$.}
\end{equation}
The proof is as follows.
The relative boundary of $\conv\{ \be_1,\ldots,\be_N \}$ can be expressed as
\[
\rbd ( \conv\{ \be_1,\ldots,\be_N \} ) = \bigcup_{i=1}^N \setF_i
\]
where
\begin{equation} \label{eq:pf3_2_t1}
\setF_i = \{ \bs \in \Rbb^N ~|~ \bs \geq \bzero, \bone^T \bs = 1, s_i = 0 \}.
\end{equation}
It follows that
\[
\bw \in \setE \cap ( \rbd ( \conv\{ \be_1,\ldots,\be_N \} ))
	\quad \Longrightarrow \quad
	\bw \in \setE \cap \setF_i \text{~for some $i \in \{1,\ldots,N\}$.}
\]
Recall $\bw_i = \tfrac{1}{N-1} \sum_{j \neq i } \be_j$.
By the Cauchy-Schwartz inequality, any $\bw \in \setF_i$ must satisfy
\[
\| \bw \| =\sqrt{N-1} \| \bw_i \| \| \bw \| \geq \sqrt{N-1} \bw_i^T \bw =  \frac{1}{\sqrt{N-1}}.
\]
Also, the above equality holds (for $\bw \in \setF_i$) if and only if $\bw= \bw_i$.
On the other hand, it can be verified that any $\bw \in \setE$ must satisfy $\| \bw \| \leq 1/\sqrt{N-1}$; see 
\eqref{eq:pf3_1_t2}.
Hence, any $\bw \in \setE \cap \setF_i$ must be given by $\bw = \bw_i$,
and applying this result to \eqref{eq:pf3_2_t1} leads to \eqref{eq:pf3_key2}.

Finally, by \eqref{eq:pf3_key1} and \eqref{eq:pf3_key2}, the desired result in  \eqref{eq:pf3_key} is obtained.

\section{An SSMF Algorithm Induced from MVIE}\label{sec:algorithm}


In this section we use the MVIE framework developed in the previous sections to derive an SSMF algorithm.

We follow the recovery procedure in Corollary~\ref{MainCorollary}, wherein the main problem is 
to solve the MVIE problem in \eqref{eq:mainP}.
To solve Problem~\eqref{eq:mainP},
we first consider DR.
The
required tool has been built in Proposition~\ref{prop:dr}:
If we can find a $2$-tuple $(\bPhi,\bb) \in \Rbb^{M \times (N-1)} \times \Rbb^M$ such that $\aff\{ \bx_1,\ldots,\bx_L \} = \setA(\bPhi,\bb)$, then the MVIE problem \eqref{eq:mainP} can be equivalently transformed to Problem \eqref{eq:drP}, restated here for convenience as follows:
\begin{equation} \label{eq:drP_rest}
\begin{aligned}
\max_{\bF', \bc'} ~ & |\det(\bF')|^2 \\
{\rm s.t.} ~ &
\setE(\bF', \bc' ) \subseteq \setX' = \conv\{ \bx_1',\ldots,\bx_L' \},
\end{aligned}
\end{equation}
where $( \bF', \bc' ) \in \Rbb^{(N-1) \times (N-1)} \times \Rbb^{N-1}$,
and $\bx_i' = \bPhi^\dag( \bx_i - \bb), i=1,\ldots,L$ are the dimensionality-reduced data points.
Specifically, recall that if $( \bF', \bc' )$ is an optimal solution to Problem~\eqref{eq:drP_rest} then $(\bF,\bc) = ( \bPhi \bF', \bPhi \bF' + \bc )$ is an optimal solution to Problem~\eqref{eq:mainP};
if $\bq' \in ( \setE( \bF', \bc' ) ) \cap ( \bdy \ \setX' )$, then $\bq = \bPhi \bq' + \bb \in ( \setE( \bF, \bc ) ) \cap ( \rbd \ \setX )$ is one of the desired contact points in \eqref{eq:main_contact_sol}.
The problem is to find one such $(\bPhi,\bb)$ from the data.
According to \cite{Chan2009}, we can extract $(\bPhi,\bb)$ from the data using affine set fitting;
it is given by $\bb= \frac{1}{L} \sum_{n=1}^L \bx_n$ and by having columns of $\bPhi$ to be first $N-1$ principal left-singular vectors of the matrix $[~ \bx_1 - \bb, \ldots, \bx_L - \bb ~]$.

Next, we show how Problem~\eqref{eq:drP_rest} can be recast as a convex problem.
To do so, we consider representing $\setX'$ in polyhedral form, that is,
\[
\setX' = \bigcap_{i=1}^K \{ \bx ~|~ \bg_i^T \bx_i \leq h_i \},
\]
for some positive integer $K$ and
for some $(\bg_i,h_i) \in \Rbb^{N-1} \times \Rbb$, $i=1,\dots,K$, with $\| \bg_i \| = 1$ without loss of generality.
Such a conversion is called facet enumeration in the literature \cite{bremner1998primal},
and in practice  $(\bg_i,h_i)_{i=1}^K$ may be obtained by calling an off-the-shelf 
algorithm such as QuickHull \cite{barber1996quickhull}.
Using the polyhedral representation of $\setX'$, Problem~\eqref{eq:drP_rest} can be reformulated as a log determinant maximization problem subject to second-order cone (SOC) constraints \cite{boyd2004convex}.
Without loss of generality, assume that $\bF'$ is symmetric and positive semidefinite.
By noting $\det(\bF') \geq 0$ and the equivalence
\begin{align}
\setE(\bF',\bc') \subseteq \bigcap_{i=1}^K \{ \bx ~|~ \bg_i^T \bx_i \leq h_i \}
    & \quad \Longleftrightarrow \quad
    \sup_{\| \balp \| \leq 1} \bg_i^T ( \bF' \balp + \bc' ) \leq h_i, ~ i=1,\ldots,K, \nonumber \\
    & \quad \Longleftrightarrow \quad
    \| \bF' \bg_i \| + \bg_i^T \bc' \leq h_i, ~i=1,\ldots,K;
    \label{eq:willneed1}
\end{align}
(see, e.g., \cite{boyd2004convex}),
Problem~\eqref{eq:drP_rest} can be rewritten as
\begin{equation} \label{eq:drP_cvx}
\begin{aligned}
\max_{\bF' \in \Sbb_+^{N-1}, \bc' \in \Rbb^{N-1}} ~ &  \log \det(\bF') \\
{\rm s.t.} ~ &
\| \bF' \bg_i \| + \bg_i^T \bc' \leq h_i, ~i=1,\ldots,K.
\end{aligned}
\end{equation}
The above problem is convex and can be readily solved by calling general-purpose convex optimization software such as CVX \cite{grant2008cvx}.
We also custom-derive a fast first-order algorithm for handling Problem~\eqref{eq:drP_cvx}.
The algorithm is described in Appendix \ref{sec:FISTA}.

The aspect of MVIE optimization is complete.
However, we should also mention how we obtain the contact points $\bq_1,\ldots,\bq_N$ in \eqref{eq:main_contact}--\eqref{eq:main_contact_sol} as they play the main role in reconstructing $\ba_1,\ldots,\ba_N$ (cf. Corollary~\ref{MainCorollary}).
It can be further shown from \eqref{eq:willneed1} that
\begin{align}
\bq' \in ( \setE(\bF',\bc') ) \cap ( \bdy \ \setX' )
    & \quad \Longleftrightarrow \quad
    \begin{aligned}
    & \bq' = \bF' \left( \frac{ \bF' \bg_i }{ \| \bF' \bg_i \| } \right) + \bc',
    ~ \| \bF' \bg_i \| + \bg_i^T \bc' = h_i, \\
   & \quad \text{for some $i=1,\ldots,K$.}
    \end{aligned}
    \label{eq:willneed2}
\end{align}
Hence, after solving Problem \eqref{eq:drP_cvx}, we can use the condition on the right-hand side of \eqref{eq:willneed2} to identify the collection of all contact points $\bq_1',\ldots,\bq_N'$.
Then, we use the relation $\bq_i = \bPhi \bq_i'+ \bb$ to construct $\bq_1,\ldots,\bq_N$.
Our
MVIE algorithm
is summarized in Algorithm~\ref{alg:MVIE}.


\begin{algorithm}[h]
\caption{An MVIE Algorithm for Blind Recovery of $\bA$}
\begin{algorithmic}[1]\label{alg:MVIE}
\STATE {\bf Given} a data matrix $\bX \in \Rbb^{M \times L}$ and a model order $N \leq \min\{M,L \}$.

\STATE Obtain the dimension-reduced data $\bx_i' = \bPhi^\dag ( \bx_i - \bb ), i=1,\dots,L$, where $(\bPhi,\bb)$ 
is
obtained by affine set fitting \cite{Chan2009}.

\STATE
Use QuickHull \cite{barber1996quickhull} or some other off-the-shelf algorithm to
enumerate the facets of
$\conv\{ \bx_1',\ldots,\bx_L'\}$,
i.e., find $(\bg_i,h_i)_{i=1}^K$ such that
$\conv\{ \bx_1',\ldots,\bx_L'\} = \cap_{i=1}^K \{ \bx ~|~ \bg_i^T \bx \leq h_i \}$.

\STATE Solve Problem~\eqref{eq:drP_cvx} either via CVX \cite{grant2008cvx} or via Algorithm \ref{alg:FISTA-huber},
and store the optimal solution obtained as $(\bF',\bc')$.

\STATE Compute the contact points $$\{ \bq_1',\ldots,\bq_N' \} = \bigg\{ \bq' = \bF' \left( \frac{ \bF' \bg_i }{ \| \bF' \bg_i \| } \right) + \bc' ~\bigg|~ \text{$i \in \{1,\ldots,K\}$ is such that $\| \bF' \bg_i \| + \bg_i^T \bc' = h_i$} \bigg\}$$

\STATE Compute the contact points $\bq_i = \bPhi \bq_i' + \bb, i=1,\ldots,N$.

\STATE Reconstruct $\ba_i=\sum_{j=1}^{N}\bq_j-(N-1)\bq_i,  i=1,\dots,N$.

\STATE {\bf Output} $\bA =[~ \ba_1,\dots,\ba_N ~]$.
\end{algorithmic}
\end{algorithm}

Some discussions are as follows.

\begin{enumerate}
\item As can be seen, the two key steps for the proposed MVIE algorithm are to perform facet enumeration and to solve a convex optimization problem.
Let us first discuss issues arising from facet enumeration.
Facet enumeration is a well-studied problem in the context of computational geometry \cite{bremner1998primal,bremner1997complexity},
and one can find off-the-shelf algorithms, such as QuickHull \cite{barber1996quickhull} and VERT2CON\footnote{\url{https://www.mathworks.com/matlabcentral/fileexchange/7895-vert2con-vertices-to-constraints}}, to perform facet enumeration.
However, it is important to note that facet enumeration is known to be {NP-hard in general} \cite{barot2017concise,boros2009generating}.
Such computational intractability was identified by finding a purposely constructed problem instance \cite{avis1997good}, which is reminiscent of the carefully constructed Klee-Minty cube for showing the worst-case complexity of the simplex method for linear programming \cite{klee1970good}.
In practice, one would argue that such worst-case instances do not happen too often.
Moreover, the facet enumeration problem is polynomial-time solvable under certain sufficient conditions,
such as the so-called ``balance condition" \cite[Theorem 3.2]{barber1996quickhull} and the case of $N=3$  \cite{cormen2001introduction}.

\item While the above discussion suggests that MVIE may not be solved in polynomial time, it is based on convex optimization and thus does not suffer from local minima.
In comparison, MVES---which enjoys the same sufficient recovery condition as MVIE---may have such issues as we will see in the numerical results in the next section.

\item We should also discuss a minor issue, namely, that of finding the contact points in Step 5 of Algorithm~\ref{alg:MVIE}.
In practice, there may be numerical errors with the MVIE 
solution,
{e.g., due to finite number of iterations or approximations involved in the algorithm.}
Also, data in reality are often noisy.
Those errors may result in identification of more than $N$ contact points as our experience suggests.
When such instances happen, we mend the problem by clustering the obtained contact points into $N$ points by standard $k$-means clustering.

\end{enumerate}

\section{Numerical Simulation and Discussion}\label{sec:simulation}

In this section we use numerical simulations to show the viability of the MVIE framework.

\subsection{Simulation Settings}

The application scenario is HU in remote sensing.
The data matrix $\bX=\bA\bS$ is synthetically generated by following the procedure in  \cite{Chan2009}.
Specifically, the columns $\ba_1,\ldots,\ba_N$ of $\bA$ are randomly selected from a library of endmember spectral signatures called the U.S. geological survey (USGS) library~\cite{USGS2007}.
The columns $\bs_1,\ldots,\bs_L$ of $\bS$ are generated by the following way:
We generate a large pool of Dirichlet distributed random vectors with concentration parameter $\bone/N$,
and then choose $\bs_1,\ldots,\bs_L$ as a subset of those random vectors whose Euclidean norms are less than or equal to a pre-specified number $r$.
{The above procedure numerically controls the pixel purity in accordance with $r$,
and therefore we will call $r$ the numerically controlled pixel purity level in the sequel.
Note that $r$ is not the uniform pixel purity level $\gamma$ in \eqref{eq:uppl},
although $r$ should closely approximate $\gamma$ when $L$ is large.
Also, we should mention that it is not feasible to control the pixel purity in accordance with $\gamma$ in our numerical experiments because verifying the value of $\gamma$ is computationally intractable~\cite{gritzmann1994complexity} (see also \cite{lin2015identifiability}).}
We set $M=224$ and $L= 1,000$.

Our main interest is to numerically verify whether the MVIE framework can indeed lead to exact recovery, and
to examine to what extent the numerical recovery results match with our theoretical claim in Theorem~\ref{thm:main}.
We measure the recovery performance by the root-mean-square (RMS) angle error
\[
\phi = \min_{ \bm \pi \in \Pi_N } \sqrt{ \frac{1}{N} \sum_{i=1}^N \left[ {\rm arccos}\left( \frac{ \bm a_i^T \hat{\bm a}_{\pi_i} }{ \| \bm a_i \| \cdot \| \hat{\bm a}_{\pi_i} \|}  \right) \right]^2  },
\]
where $\Pi_N$ denotes the set of all permutations of $\{1,\ldots,N\}$,
and $\hat{\bA}$ denotes an estimate of $\bA$ by an algorithm.
We use $200$ independently generated realizations to evaluate the average RMS angle errors.
Two versions of the MVIE implementations in Algorithm~\ref{alg:MVIE} are considered.
The first calls the general-purpose convex optimization software CVX as to solve the MVIE problem,
while the second applies the custom-derived algorithm in Algorithm \ref{alg:FISTA-huber} (with $\rho= 150$, $\epsilon= 2.22 \times 10^{-16}$, {$\alpha= 2$,} $\beta=0.6$) to solve the MVIE problem (approximately).
For convenience, the former and latter will be called ``MVIE-CVX'' and ``MVIE-FPGM'', resp.
We also tested some other algorithms for benchmarking,
namely, the successive projection algorithm (SPA) \cite{gillis2014fast}, SISAL \cite{Dias2009}
and MVES \cite{Chan2009}.
SPA is a fast pure-pixel search, or separable NMF, algorithm.
SISAL and MVES are non-convex optimization-based algorithms under the MVES framework.
{Following the original works, we initialize SISAL by vertex component analysis (a pure-pixel search algorithm) \cite{nascimento2005vertex} and initialize MVES by the solution of a convex feasibility problem \cite[Problem (43)]{Chan2009}.}
All the algorithms are implemented under Mathworks Matlab R2015a,
and they were run on a computer with Core-i7-4790K CPU (3.6 GHz CPU speed) and with 16GB RAM.

\subsection{Recovery Performance}

\begin{figure}[H]
	\centering
	\includegraphics[width=\linewidth]{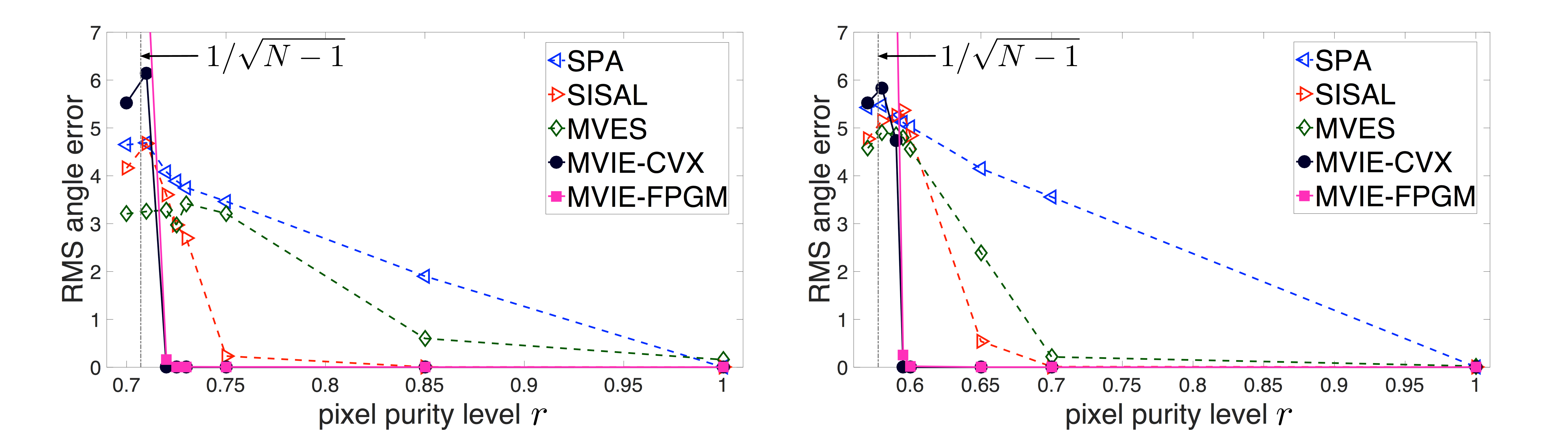}\\
	(a) $N=3$\hfil\hfil (b) $N=4$
	\includegraphics[width=\linewidth]{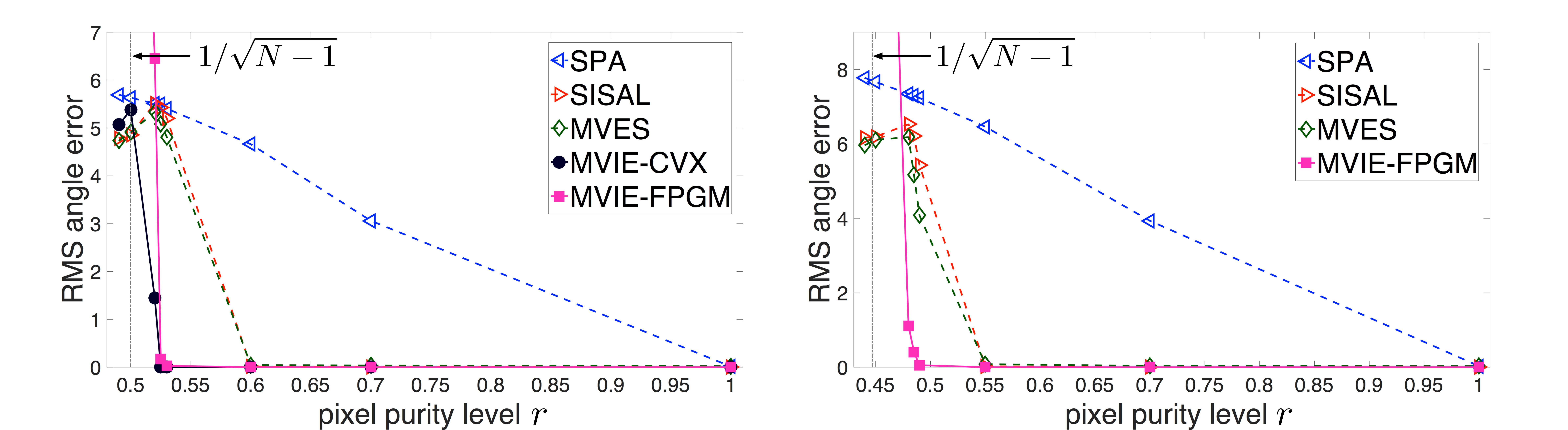}\\
	(c) $N=5$\hfil\hfil (d) $N=6$
	\includegraphics[width=\linewidth]{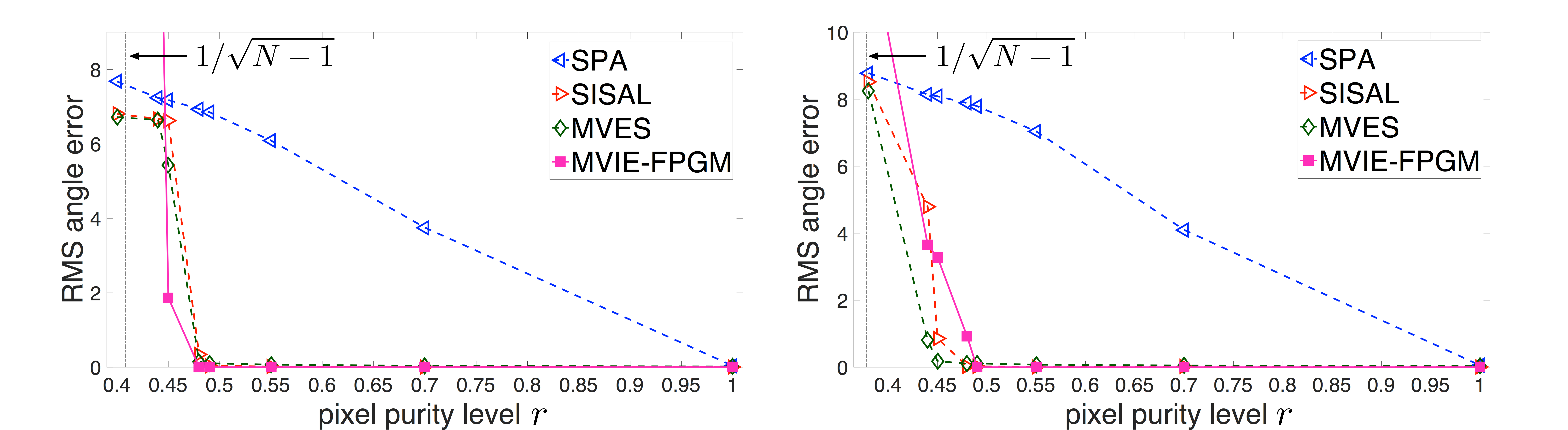}\\
	{(e) $N=7$\hfil\hfil (f) $N=8$}
	\caption{Recovery performance of the SSMF algorithms with respect to the numerically controlled pixel purity level $r$. $M = 224,L = 1,000$, the noiseless case.}
	\label{fig:sim1}
\end{figure}

Figure \ref{fig:sim1} plots the average RMS angle errors of the various algorithms versus the (numerically controlled) pixel purity level $r$.
{As a supplementary result for Figure \ref{fig:sim1}, the precise values of the averages and standard deviations of the RMS angle errors are further shown in Table~\ref{tab:sim1}.
Let us first examine the cases of $3 \leq N \leq 5$.
MVIE-CVX achieves essentially perfect recovery performance when the pixel purity level $r$ is larger than $1/\sqrt{N-1}$ by a margin of $0.025$.
This corroborates our sufficient recovery condition in Theorem~\ref{thm:main}.
We also see from Figure \ref{fig:sim1} that MVIE-FPGM has similar performance trends.
However, upon a closer look at the numbers in Table~\ref{tab:sim1}, MVIE-FPGM is seen to have slightly higher RMS angle errors than MVIE-CVX.
This is because MVIE-FPGM employs an approximate solver for the MVIE problem (Algorithm \ref{alg:FISTA-huber}) to trade for better runtime;} 
the runtime performance will be illustrated later.

\begin{figure}[h!]
	\centering
	\includegraphics[width=\linewidth]{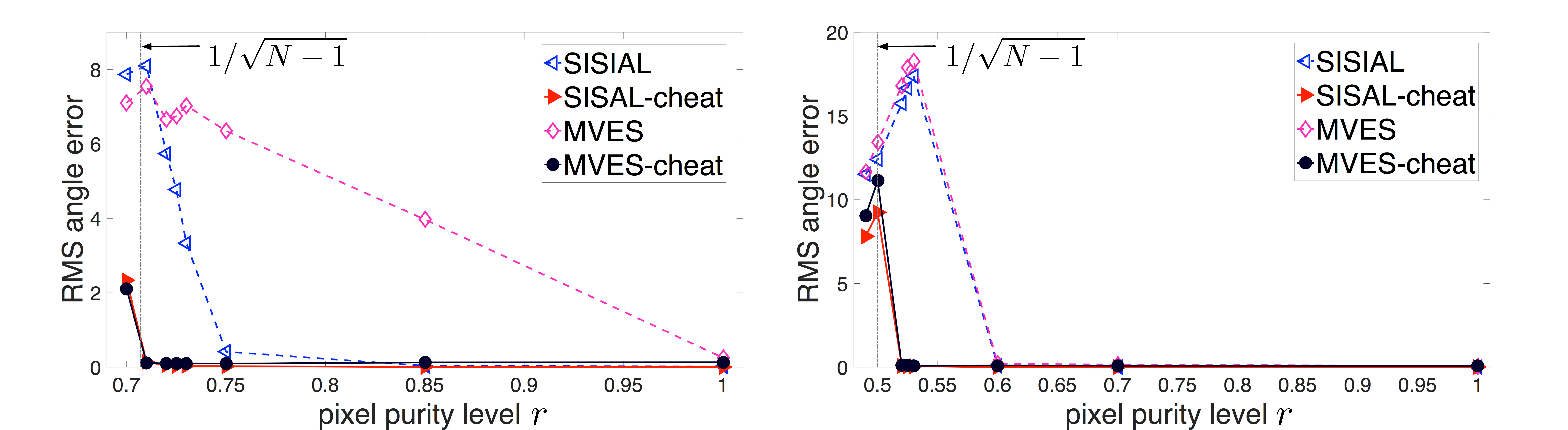}
	(a) $N=3$\hfil\hfil (b) $N=5$
	\caption{Recovery performance of MVES and SISAL under different initializations.}
	\label{fig:sim15}
\end{figure}

Let us also compare the MVIE algorithms and the other benchmarked algorithms, again, for $3 \leq N \leq 5$.
SPA has its recovery performance deteriorating as the pixel purity level $r$ decreases.
This is expected as separable NMF or pure-pixel search is based on the separability or pure-pixel assumption, which corresponds to $r=1$ in our simulations
{(with high probability)}.
SISAL and MVES, on the other hand, are seen to give perfect recovery for a range of values of $r$.
However, when we observe the transition points from perfect recovery to imperfect recovery, SISAL and MVES appear not as resistant to lower pixel purity levels as MVIE-CVX and MVIE-FPGM.
The main reason of this is that SISAL and MVES can suffer from convergence to 
local minima.
To support our argument, Figure~\ref{fig:sim15} gives an additional numerical result where we use slightly perturbed versions of the groundtruth $\ba_1,\ldots,\ba_N$ as the initialization and see if MVES and SISAL would converge to a different solution.
``SISAL-cheat'' and ``MVES-cheat'' refer to MVES and SISAL run under such cheat initializations, resp.;
``SISAL'' and ``MVES'' refer to the original SISAL and MVES.
We see from Figure~\ref{fig:sim15} that the two can have significant gaps, which verifies that SISAL and MVES can be sensitive to initializations.

\begin{table}[hbt]
	\centering
	{\caption{RMS angle error (deg.) of the various algorithms. The simulation settings are the same as those in Figure~\ref{fig:sim1}.}
	\label{tab:sim1}
	\begin{tabular}{|c|c|c|c|c|c|c|}\hline
		$N$ & $r$ & SPA & SISAL & MVES & MVIE-CVX & MVIE-FPGM \\\hline
		\multirow{3}{*}{3} & 0.72 & 4.081$\pm$0.538 & 3.601$\pm$2.270 & 3.286$\pm$2.433 & 0.001$\pm$0.001 & 0.161$\pm$0.376 \\
		& 0.85 & 1.903$\pm$0.121 & 0.006$\pm$0.003 & 0.602$\pm$0.638 & 0.000$\pm$0.000 & 0.003$\pm$0.002 \\
		& 1 & 0.002$\pm$0.001 & 0.003$\pm$0.001 & 0.158$\pm$0.324 & 0.000$\pm$0.000 & 0.002$\pm$0.002 \\\hline
		\multirow{3}{*}{4} & 0.595 & 5.114$\pm$0.389 & 5.369$\pm$1.147 & 4.800$\pm$1.984 & 0.006$\pm$0.011 & 0.257$\pm$0.251 \\
		& 0.7 & 3.558$\pm$0.318 & 0.012$\pm$0.007 & 0.216$\pm$0.297 & 0.000$\pm$0.000 & 0.002$\pm$0.001 \\
		& 1 & 0.007$\pm$0.004 & 0.003$\pm$0.001 & 0.023$\pm$0.042 & 0.000$\pm$0.000 & 0.002$\pm$0.001 \\\hline
		\multirow{3}{*}{5} & 0.525 & 5.494$\pm$0.210 & 5.422$\pm$0.973 & 5.082$\pm$1.485 & 0.004$\pm$0.009 & 0.169$\pm$0.174 \\
		& 0.7 & 3.061$\pm$0.150 & 0.007$\pm$0.005 & 0.036$\pm$0.046 & 0.000$\pm$0.000 & 0.002$\pm$0.000 \\
		& 1 & 0.014$\pm$0.007 & 0.002$\pm$0.001 & 0.024$\pm$0.037 & 0.000$\pm$0.000 & 0.002$\pm$0.000 \\\hline
		\multirow{3}{*}{6} & 0.48 & 7.343$\pm$0.232 & 6.526$\pm$1.166 & 6.180$\pm$1.875 & - & 1.117$\pm$1.629 \\
		& 0.7 & 3.935$\pm$0.193 & 0.008$\pm$0.006 & 0.036$\pm$0.041 & - & 0.001$\pm$0.000 \\
		& 1 & 0.030$\pm$0.014 & 0.004$\pm$0.001 & 0.031$\pm$0.045 & - & 0.002$\pm$0.000 \\\hline
		\multirow{3}{*}{7} & 0.45 & 7.178$\pm$0.193 & 6.629$\pm$1.255 & 5.438$\pm$2.883 & - & 1.868$\pm$2.355 \\
		& 0.7 & 3.752$\pm$0.210 & 0.011$\pm$0.018 & 0.038$\pm$0.040 & - & 0.001$\pm$0.000 \\
		& 1 & 0.040$\pm$0.019 & 0.004$\pm$0.001 & 0.020$\pm$0.029 & - & 0.001$\pm$0.000 \\\hline
		\multirow{3}{*}{8} & 0.44 & 8.140$\pm$0.257 & 4.791$\pm$3.108 & 0.802$\pm$1.806 & - & 3.659$\pm$1.768 \\
		& 0.7 & 4.099$\pm$0.271 & 0.019$\pm$0.057 & 0.052$\pm$0.053 & - & 0.001$\pm$0.000 \\
		& 1 & 0.055$\pm$0.023 & 0.005$\pm$0.001 & 0.034$\pm$0.048 & - & 0.001$\pm$0.000\\\hline
	\end{tabular}
}
\end{table}

Next, we examine the cases of $6 \leq N \leq 8$ in Figure \ref{fig:sim1}.
For these cases we did not test MVIE-CVX because it runs slowly for large $N$.
By comparing the transition points from perfect recovery to imperfect recovery,
we observe that MVIE-FPGM is better than SISAL and MVES for $N=6$,
on a par with SISAL and MVES for $N=7$,
and worse than SISAL and MVES for $N=8$;
the gaps are nevertheless not significant.

%

\begin{figure}[h]
	\centering
	{	\includegraphics[width=1.04\linewidth]{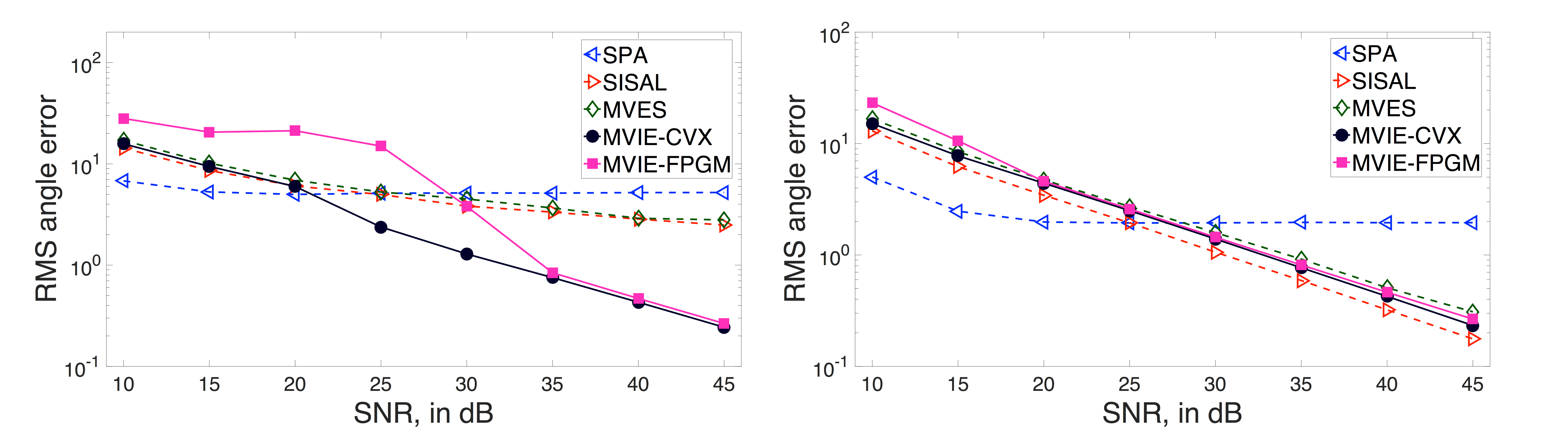}\\
	(a) \quad$r=0.55$\hfil\hfil (b) $r=0.8$
	\caption{Recovery performance of the SSMF algorithms with respect to the SNR. $M = 224, N=5, L = 1,000$.}
	}
	\label{fig:sim2}
\end{figure}

The MVIE framework we established assumes the noiseless case.
Having said so, it is still interesting to evaluate how MVIE performs in the noisy case.
Figure~\ref{fig:sim2} plots the RMS angle error performance of the various algorithms versus the signal-to-noise ratio (SNR), with $N= 5$.
Specifically, we add independent and identically distributed mean-zero Gaussian noise to the data, and
{the SNR is defined as ${\rm SNR} = ( \sum_{i=1}^L \|{\bf x}_i\|^2 )/( \sigma^2ML )$ where $\sigma^2$ is the noise variance.}
We observe that MVIE-CVX performs better than SISAL and MVES when $r= 0.55$ and ${\rm SNR} \geq 25$dB; MVIE-FPGM does not work as good as MVIE-CVX but still performs better than SISAL and MVES when $r= 0.55$ and ${\rm SNR} \geq 35$dB.
This suggests that MVIE may work better for lower pixel purity levels.

%
%
%

\subsection{Runtime Performance}

We now turn our attention to runtime performance.
Table~\ref{table:time1} shows the runtimes of the various algorithms for various $N$ and $r$.
Our observations are as follows.
First, we see that MVIE-CVX is slow especially for larger $N$.
{The reason is that CVX calls an interior-point algorithm to solve the MVIE problem,
and second-order methods such as interior-point methods are known to be less efficient when dealing with problems with many constraints.}
Second, MVIE-FPGM, which uses an approximate MVIE solver based on first-order methodology, runs much faster than MVIE-CVX.
Third, MVIE-FPGM is faster than MVES for $N \leq 7$ and SISAL for $N \leq 4$, but is slower than the latters otherwise.

\begin{table}[H]
	\centering
	{
	\caption{Runtimes (sec.) of the various algorithms. The simulation settings are the same as those in Figure~\ref{fig:sim1}.}
	\label{table:time1}
	\begin{tabular}{|c|c|c|c|c|c|c|}\hline
		$N$ & $r$ & SPA & SISAL & MVES & MVIE-CVX & MVIE-FPGM \\\hline
		\multirow{3}{*}{3} & 0.72 & 0.008$\pm$0.008 & 0.288$\pm$0.011 & 0.285$\pm$0.244 & 0.613$\pm$0.044 & 0.031$\pm$0.016 \\
		& 0.85 & 0.008$\pm$0.008 & 0.282$\pm$0.009 & 0.803$\pm$0.569 & 0.466$\pm$0.039 & 0.034$\pm$0.028 \\
		& 1 & 0.006$\pm$0.008 & 0.273$\pm$0.009 & 1.506$\pm$0.848 & 0.314$\pm$0.034 & 0.041$\pm$0.031 \\\hline
		\multirow{3}{*}{4} & 0.595 & 0.009$\pm$0.008 & 0.323$\pm$0.010 & 0.766$\pm$0.759 & 4.112$\pm$0.213 & 0.106$\pm$0.048 \\
		& 0.7 & 0.009$\pm$0.008 & 0.316$\pm$0.010 & 3.327$\pm$1.593 & 3.579$\pm$0.202 & 0.042$\pm$0.019 \\
		& 1 & 0.006$\pm$0.008 & 0.301$\pm$0.009 & 5.305$\pm$1.015 & 1.378$\pm$0.176 & 0.046$\pm$0.040 \\\hline
		\multirow{3}{*}{5} & 0.525 & 0.010$\pm$0.008 & 0.371$\pm$0.009 & 2.228$\pm$1.825 & 33.115$\pm$2.362 & 0.514$\pm$0.105 \\
		& 0.7 & 0.012$\pm$0.005 & 0.359$\pm$0.009 & 10.528$\pm$1.955 & 32.642$\pm$3.149 & 0.441$\pm$0.180 \\
		& 1 & 0.009$\pm$0.004 & 0.339$\pm$0.008 & 11.859$\pm$1.185 & 10.012$\pm$1.651 & 0.340$\pm$0.071 \\\hline
		\multirow{3}{*}{6} & 0.48 & 0.016$\pm$0.003 & 0.444$\pm$0.010 & 5.303$\pm$3.920 & - & 2.354$\pm$0.150 \\
		& 0.7 & 0.014$\pm$0.007 & 0.396$\pm$0.009 & 19.825$\pm$1.737 & - & 2.229$\pm$0.321 \\
		& 1 & 0.009$\pm$0.008 & 0.371$\pm$0.008 & 20.033$\pm$1.973 & - & 1.220$\pm$0.130 \\\hline
		\multirow{3}{*}{7} & 0.45 & 0.018$\pm$0.007 & 0.489$\pm$0.013 & 11.504$\pm$6.392 & - & 10.648$\pm$1.113 \\
		& 0.7 & 0.017$\pm$0.005 & 0.426$\pm$0.011 & 33.706$\pm$1.946 & - & 19.331$\pm$0.830 \\
		& 1 & 0.011$\pm$0.009 & 0.402$\pm$0.009 & 34.006$\pm$2.790 & - & 7.321$\pm$0.876 \\\hline
		\multirow{3}{*}{8} & 0.44 & 0.021$\pm$0.008 & 0.549$\pm$0.021 & 32.663$\pm$6.465 & - & 77.600$\pm$8.446 \\
		& 0.7 & 0.023$\pm$0.008 & 0.468$\pm$0.012 & 67.577$\pm$2.001 & - & 157.313$\pm$5.637 \\
		& 1 & 0.015$\pm$0.010 & 0.435$\pm$0.010 & 60.882$\pm$4.502 & - & 57.613$\pm$8.386\\\hline
	\end{tabular}
	}
\end{table}

~ \\

In the previous section we discussed the computational bottleneck of facet enumeration in MVIE.
To get some ideas on the situation in practice,
we show the runtime breakdown of MVIE-FPGM in Table~\ref{table:time2}.
We see that {facet enumeration takes only about $10\%$ to $33\%$ of the total runtime in MVIE-FPGM.}
{But there is a caveat:
	Facet enumeration can output a large number of facets $K$, and from Table~\ref{table:time2} we observe that this is particularly true when $N$ increases.
	Since $K$ is the number of SOC constraints of the MVIE problem \eqref{eq:drP_cvx},
	solving the MVIE problem  for larger $N$ becomes more difficult computationally.}
While 
the main contribution of this paper is to introduce a new theoretical SSMF framework through MVIE,
as a future direction it would be interesting to study how 
the aforementioned issue can be mitigated.

\begin{table}[h]
	\centering
	{
	\caption{Detailed runtimes (sec.) of MVIE-FPGM. The simulation settings are the same as those in Figure~\ref{fig:sim1}.}
	\label{table:time2}
	\begin{tabular}{|c|c||c|c|c||c|}\hline
		\multirow{2}{*}{$N$} & \multirow{2}{*}{$r$} & \multicolumn{3}{c||}{Runtime} & Number of facets $K$  \\\cline{3-5}
		&  & MVIE-FPGM & Facet enumeration & FPGM+Others & by facet enumeration \\\hline
		\multirow{3}{*}{3} & 0.72 & 0.031$\pm$0.016 & 0.007$\pm$0.002 & 0.024$\pm$0.014 & 44.03$\pm$3.48 \\
		& 0.85 & 0.034$\pm$0.028 & 0.007$\pm$0.002 & 0.027$\pm$0.026 & 29.91$\pm$3.98 \\
		& 1 & 0.041$\pm$0.031 & 0.007$\pm$0.002 & 0.035$\pm$0.030 & 16.12$\pm$3.12 \\\hline
		\multirow{3}{*}{4} & 0.595 & 0.106$\pm$0.048 & 0.022$\pm$0.005 & 0.084$\pm$0.043 & 365.68$\pm$17.64 \\
		& 0.7 & 0.042$\pm$0.019 & 0.020$\pm$0.003 & 0.022$\pm$0.016 & 318.01$\pm$18.26 \\
		& 1 & 0.046$\pm$0.040 & 0.012$\pm$0.004 & 0.034$\pm$0.035 & 114.62$\pm$18.49 \\\hline
		\multirow{3}{*}{5} & 0.525 & 0.514$\pm$0.105 & 0.109$\pm$0.006 & 0.405$\pm$0.100 & 2208.76$\pm$101.54 \\
		& 0.7 & 0.441$\pm$0.180 & 0.112$\pm$0.005 & 0.329$\pm$0.174 & 2055.93$\pm$88.57 \\
		& 1 & 0.340$\pm$0.071 & 0.052$\pm$0.006 & 0.288$\pm$0.065 & 764.00$\pm$102.10 \\\hline
		\multirow{3}{*}{6} & 0.48 & 2.354$\pm$0.150 & 0.663$\pm$0.039 & 1.691$\pm$0.111 & 11901.32$\pm$699.30 \\
		& 0.7 & 2.229$\pm$0.321 & 0.760$\pm$0.028 & 1.469$\pm$0.293 & 13064.35$\pm$511.29 \\
		& 1 & 1.220$\pm$0.130 & 0.345$\pm$0.036 & 0.875$\pm$0.094 & 4982.35$\pm$611.11 \\\hline
		\multirow{3}{*}{7} & 0.45 & 10.648$\pm$1.113 & 2.906$\pm$0.311 & 7.742$\pm$0.801 & 49377.95$\pm$4454.29 \\
		& 0.7 & 19.331$\pm$0.830 & 5.947$\pm$0.211 & 13.384$\pm$0.619 & 81631.50$\pm$3398.41 \\
		& 1 & 7.321$\pm$0.876 & 2.541$\pm$0.268 & 4.780$\pm$0.608 & 29448.52$\pm$4109.01 \\\hline
		\multirow{3}{*}{8} & 0.44 & 77.600$\pm$8.446 & 19.226$\pm$2.171 & 58.374$\pm$6.276 & 279720.40$\pm$29481.38 \\
		& 0.7 & 157.313$\pm$5.637 & 51.648$\pm$1.772 & 105.665$\pm$3.865 & 495624.59$\pm$18868.73 \\
		& 1 & 57.613$\pm$8.386 & 22.914$\pm$3.042 & 34.700$\pm$5.344 & 161533.59$\pm$24957.12\\\hline
	\end{tabular}
	}
\end{table}

\section{Conclusion}\label{sec:conclusion}

In this paper we have established a new SSMF framework through analyzing an MVIE problem.
As the {main contribution}, we showed that the MVIE framework can admit exact recovery beyond separable or pure-pixel problem instances, and that its exact recovery condition is as good as that of the MVES framework.
However, unlike MVES which requires one to solve a non-convex problem, the MVIE framework suggests a two-step solution, namely, facet enumeration and convex optimization.
The viability of the MVIE framework was shown by numerical results, and it was illustrated that MVIE exhibits stable performance over a wide range of pixel purity levels.
%
Furthermore, we should mention three open questions arising from the current investigation:
\begin{itemize}
	\item {How can we make facet enumeration more efficient in the sense of generating less facets, thereby improving the efficiency of computing the MVIE?
	In this direction it is worthwhile to point out the subset-separable NMF work \cite{ge2015intersecting} which considers a similar facet identification problem but operates on rather different sufficient recovery conditions.}
	
	\item How can we 
			handle the MVIE computations efficiently
			when
			the number of facets, even with a better facet enumeration procedure, is still very large? 
		One possibility is to consider the active set strategy, which was found to be very effective in dealing with the minimum volume covering ellipsoid (MVCE) problem \cite{sun2004computation,gillis2015semidefinite}.
		While the MVCE problem is not identical to the MVIE problem, it will be interesting to investigate how the insights in the aforementioned references can be used in our problem at hand.
		
	\item {How should we modify the MVIE formulation in the noisy case such that it may offer better robustness to noise---both practically and provably?}
\end{itemize}
We hope this new framework might inspire more theoretical and practical results in tackling SSMF.



\appendix

\section{Proof of Proposition \ref{prop:dr}}\label{sec:proof-prop1}

We will use the following results.
\begin{Fact} \label{fac:simple} Let $f(\balp)= \bPhi \balp + \bb$ where $(\bPhi, \bb) \in \Rbb^{m \times n} \times  \Rbb^m$ and $\bPhi$ has full column rank.
The following results hold.
\begin{enumerate}[\leftmargin=1cm]
\item (a) Let $\setC$ be a non-empty set in $\Rbb^m$ with $\setC \subseteq \setA(\bPhi,\bb)$.
Then $$\rbd(f^{-1}(\setC)) = f^{-1}( \rbd \ \setC ).$$

\smallskip

\item (b) Let $\setC_1, \setC_2$ be sets in $\Rbb^m$ with $\setC_1, \setC_2 \subseteq \setA(\bPhi,\bb)$. Then
\[
\setC_1 \subseteq \setC_2  \quad \Longleftrightarrow \quad  f^{-1}(\setC_1) \subseteq f^{-1}(\setC_2).
\]


\end{enumerate}
\end{Fact}
The results in the above fact
may be easily deduced or found in textbooks.

First, we prove the feasibility results in Statements (a)--(b) of Proposition \ref{prop:dr}.
Let $(\bF,\bc)$ be a feasible solution to Problem~\eqref{eq:mainP}. Since
\[
\setE(\bF,\bc) \subseteq \setX \subseteq \aff\{ \bx_1,\ldots,\bx_L \} = \setA(\bPhi,\bb),
\]
it holds that
\[
\bff_i + \bc = \bPhi \balp_i + \bb, ~ i=1,\ldots,N, \qquad \bc = \bPhi \bc' + \bb,
\]
for some $\balp_1,\ldots,\balp_N, \bc' \in \Rbb^{N-1}.$
By letting $\bff_i' = \balp_i - \bc', i=1,\ldots,N$,
one can show that $\bF' = [~ \bff_1',\ldots, \bff_N' ~]$ and $\bc'$ are uniquely given by
$(\bF',\bc') = ( \bPhi^\dag \bF, \bPhi^\dag(\bc- \bb))$.
Also, by letting $f(\balp)= \bPhi \balp + \bb$, it can be verified that
\[
f^{-1}( \setE(\bF,\bc) ) = \setE( \bF', \bc' ).
\]
Similarly, for $\setX$, we have $\bx_i \in \setX \subseteq \setA(\bPhi,\bb)$.
This means that $\bx_i$ can be expressed as $\bx_i = \bPhi \bx_i' + \bb$ for some $\bx_i' \in \Rbb^{N-1}$, and it can be verified that $\bx_i'$ is uniquely given by $\bx_i' = \bPhi^\dag( \bx_i - \bb)$.
Subsequently it can be further verified that
\[
f^{-1}( \setX ) = \setX'.
\]
Hence, by using Fact~\ref{fac:simple}.(b) via setting $\setC_1 = \setE(\bF,\bc), \setC_2 = \setX$, we get $\setE( \bF', \bc' ) \subseteq \setX'$.
Thus, $(\bF',\bc')$ is a feasible solution to Problem~\eqref{eq:drP}, and we have proven the feasibility result in Statement (a) of Proposition \ref{prop:dr}.
The proof of the feasibility result in Statement (b) of Proposition \ref{prop:dr} follows the same proof method, and we omit it for brevity.

Second, we prove the optimality results in Statements (a)--(b) of Proposition \ref{prop:dr}.
Let $(\bF,\bc)$ be an optimal solution to Problem~\eqref{eq:mainP},
$(\bF',\bc')$ be equal to $( \bPhi^\dag \bF, \bPhi^\dag(\bc- \bb))$ which is feasible to Problem~\eqref{eq:drP},
and $v_{\rm opt}$ be the optimal value of Problem~\eqref{eq:mainP}.
Then we have
\[
v_{\rm opt} = \det( \bF^T \bF ) = \det( (\bF')^T \bPhi^T \bPhi \bF ) = | \det(\bF') |^2 \det(\bPhi^T \bPhi) \geq v_{\rm opt}' \det(\bPhi^T \bPhi),
\]
where $v_{\rm opt}'$ denotes the optimal value of Problem~\eqref{eq:drP}.
Conversely, by redefining $(\bF',\bc')$ as an optimal solution to Problem~\eqref{eq:drP} and $(\bF,\bc)= (\bPhi\bF',\bPhi\bc'+ \bb)$ (which is feasible to Problem~\eqref{eq:mainP}), we also get
\[
v_{\rm opt}' = | \det(\bF') |^2 = \frac{1}{\det(\bPhi^T \bPhi)} \det( \bF^T \bF ) \geq \frac{1}{\det(\bPhi^T \bPhi)}v_{\rm opt}.
\]
The above two equations imply $v_{\rm opt} = v_{\rm opt}' \det(\bPhi^T \bPhi)$,
and it follows that the optimal solution results in Statements (a)--(b) of Proposition \ref{prop:dr} are true.

Third, we prove Statement (c) of Proposition \ref{prop:dr}.
Recall from \eqref{eq:dimX} that $\dim \setX = N-1$ (also recall that the result is based on the premise of (A2)--(A3)).
From the development above, one can show that
\[
\setX = \{ \bPhi \bx' + \bb ~|~ \bx' \in \setX' \}.
\]
It can be further verified from the above equation and the full column rank property of $\bPhi$ that $\dim \setX' = \dim \setX = N-1$ must hold.
In addition, as a basic convex analysis result, a convex set $\setC$ in $\Rbb^m$ has non-empty interior if $\dim \setC = m$.
This leads us to the conclusion that $\setX'$ has non-empty interior.

Finally, we prove Statement (d) of Proposition \ref{prop:dr}.
The results therein are merely applications of Fact~\ref{fac:simple};
e.g., $\setC_1 = \{ \bq \}, \setC_2 = \setE$ for $\bq \in \setE \Longrightarrow \bq' \in \setE'$,
$\setC_1 = \{ \bq \}, \setC_2 = \rbd \ \setX$ for $\bq \in \rbd \ \setX \Longrightarrow \bq' \in \rbd ( f^{-1}(\setX) ) = \bdy \ \setX'$,
and so forth.

\section{Fast Proximal Gradient Algorithm for Handling Problem \eqref{eq:drP_cvx}}\label{sec:FISTA}

In this appendix we derive a fast algorithm for handling the MVIE problem in \eqref{eq:drP_cvx}.
Let us describe the formulation used.
Instead of solving Problem~\eqref{eq:drP_cvx} directly, we employ an approximate formulation as follows
\begin{equation} \label{eq:drP_cvx_penalized}
\begin{aligned}
\min_{\bF' \in \Sbb_+^{N-1}, \bc' \in \Rbb^{N-1}} ~ &  -\log \det(\bF')  + \rho \sum_{i=1}^K \psi(
\| \bF' \bg_i \| + \bg_i^T \bc' - h_i ),
\end{aligned}
\end{equation}
for a pre-specified constant $\rho > 0$ and for some convex differentiable function $\psi: \Rbb \rightarrow \Rbb$ such that $\psi(x) = 0$ for $x \leq 0$ and $\psi(x) > 0$ for $x > 0$;
specifically our choice of $\psi$ is the one-sided Huber function, i.e.,
\[
\psi(z)=
\begin{cases}
0,&~z< 0,
\\
\frac{1}{2}z^2,&~0\leq z\leq 1,
\\
z-\frac{1}{2},&~z>1.
\end{cases}
\]
Our approach is to use a penalized, or ``soft-constrained'', convex formulation in place of Problem~\eqref{eq:drP_cvx},
whose SOC constraints may not be easy
{to deal with as ``hard constraints''.}
Problem \eqref{eq:drP_cvx_penalized} has a nondifferentiable and unbounded-above objective function.
To facilitate our algorithm design efforts later, we further approximate the problem by
\begin{equation} \label{eq:drP_cvx_penalized2}
\begin{aligned}
\min_{\bF' \in \mathcal{W}, \bc' \in \Rbb^{N-1}} ~ &  - \log \det(\bF')  + \rho \sum_{i=1}^K \psi(
\sqrt{ \| \bF' \bg_i \|^2 + \epsilon } + \bg_i^T \bc' - h_i ),
\end{aligned}
\end{equation}
for some small constant $\epsilon > 0$, where $\mathcal{W}\triangleq\{
\bW\in\mathbb{S}^{N-1} \mid\lambda_\text{min}(\bW)\geq\epsilon
\}$.

Now we describe the algorithm.
We employ the fast proximal gradient method (FPGM) or FISTA \cite{beck2009fast},
which is known to guarantee a convergence rate of $\mathcal{O}(1/k^2)$ under certain premises; here, $k$ is the iteration number.
For notational convenience, let us denote $n=N-1$, $\bW=\bF'$, $\by=\bc'$, and rewrite Problem \eqref{eq:drP_cvx_penalized2} as
\begin{align}
\label{formulation}
	\min_{\substack{\bW\in\mathbb{R}^{n\times n}\\\by\in\Rbb^{n}}}\underbrace{\sum_{i=1}^K \psi\left(
	\sqrt{\|\bW\bg_i\|^2+\epsilon}+\bg_i^T\by-h_i
	\right)}_{\triangleq f(\bW,\by)}+
	\underbrace{{I}_\mathcal{W}(\bW)-\frac{1}{\rho}\log\det(\bW)}_{\triangleq g(\bW)},
\end{align}
where $I_{\mathcal{W}}(\cdot)$ is the indicator function of $\mathcal{W}$.
By applying FPGM to the formulation in \eqref{formulation}, we obtain Algorithm \ref{alg:FISTA-huber}.
In the algorithm, the notation $\langle \cdot, \cdot \rangle$ stands for the inner product, $\| \cdot \|$ still stands for the Euclidean norm, $\psi'$ is the differentiation of $\psi$, and ${\rm prox}_{f}(\bz) = \arg \min_{\bx} \frac{1}{2} \| \bz - \bx \|^2 + f(\bx)$ is the proximal mapping of $f$.
The algorithm requires computations of the proximal mapping ${\rm prox}_{tg}(\bW - t \nabla_\bW f)$.
The solution to our proximal mapping is described in the following fact.
\begin{Fact} \label{fact:pm}
Consider the proximal mapping ${\rm prox}_{tg}(\bV)$ where the function $g$ has been defined in \eqref{formulation} and $t > 0$.
Let $\bV_{\rm sym} = \frac{1}{2} ( \bV + \bV^T )$, and let $\bV_{\rm sym} = \bU \bLam \bU^T$ be the symmetric eigendecomposition of $\bV_{\rm sym}$ where $\bU \in \Rbb^{n \times n}$ is orthogonal and $\bLam \in \Rbb^{n \times n}$ is diagonal with diagonal elements given by $\lambda_1,\ldots,\lambda_n$.
We have
\[
{\rm prox}_{tg}(\bV) = \bU \bD \bU^T
\]
where $\bD \in \Rbb^{n \times n}$ is diagonal with diagonal elements given by $d_i = \max\left\{\frac{\lambda_i+\sqrt{\lambda_i^2+4t/\rho}}{2},\epsilon\right\}$, $i=1,\ldots,n$.
\end{Fact}
The proof of the above fact will be given in Appendix~\ref{app:fact:pm}.
%
Furthermore, we should mention convergence.
FPGM is known to have a $\mathcal{O}(1/k^2)$ convergence rate if the problem is convex and $f$ has a Lipschitz continuous gradient.
In Appendix \ref{app:Lip}, we show that $f$ has a Lipschitz continuous gradient.

\begin{algorithm}[h]
\caption{FPGM for Solving Problem \eqref{formulation}}
\begin{algorithmic}[1]\label{alg:FISTA-huber}
\STATE {\bf Given} $\epsilon>0$, $\rho>0$, $(\bm g_i,h_i)_{i=1}^K$, $t_{\rm max}>0$,
{$\alpha \geq 1$,} $\beta\in(0,1)$,
and a starting point $(\bW,\by) \in \mathcal{W} \times \Rbb^n$.


\STATE Set $k:=1$, $u_0 = 0$, $(\bW^0,\by^0) = (\bW,\by)$, {$t:= t_{\rm max}$.}

\REPEAT


\STATE $\nabla_\bW f := \sum_{i=1}^{K}
\frac{\psi'(\sqrt{\|\bW\bg_i\|^2+\epsilon}+\bg_i^T\by-h_i)}{\sqrt{\|\bW\bg_i\|^2+\epsilon}}(\bW\bg_i\bg_i^T)$;

\STATE
$\nabla_\by f := \sum_{i=1}^{K}
{\psi'(\sqrt{\|\bW\bg_i\|^2+\epsilon}+\bg_i^T\by-h_i)}~\!\bg_i$;



\STATE
{$t:= \alpha t$;}

\STATE
$\bW^k := {\rm prox}_{tg} (\bW - t \nabla_\bW f )$, $\by^k := \by - t \nabla_\by f$;

\STATE
{\em \% line search}

\WHILE{$f(\bW^k,\by^k)>f(\bW,\by)+
\langle
(\nabla_\bW f,\nabla_\by f),
(\bW^k,\by^k)-(\bW,\by)
\rangle
+\frac{1}{2t}\|(\bW^k,\by^k)-(\bW,\by)\|^2$}

\STATE $t:=\beta t$;

\STATE
$\bW^k := {\rm prox}_{tg} (\bW - t \nabla_\bW f )$, $\by^k := \by - t \nabla_\by f$;

\ENDWHILE

\STATE $u_{k}=\frac{1}{2}\left(1+\sqrt{1+4u_{k-1}^2}\right)$;

\STATE
 $(\bW,\by):=
(\bW^{k},\by^{k})+
\frac{u_{k-1}-1}{u_{k}}\left(
(\bW^{k},\by^{k})-
(\bW^{{k-1}},\by^{{k-1}})
\right)$;

\STATE $k:=k+1$;

\UNTIL a pre-specified stopping rule is satisfied.

\STATE {\bf Output} $(\bW^{k-1},\by^{k-1})$.
\end{algorithmic}
\end{algorithm}

\subsection{Proof of Fact~\ref{fact:pm}} \label{app:fact:pm}
It can be verified that for any symmetric $\bW$, we have $\| \bV - \bW \|^2 = \| \bV_{\rm sym} - \bW \|^2 + \| \frac{1}{2} (\bV - \bV^T) \|^2$.
Thus, the proximal mapping ${\rm prox}_{tg}(\bV)$ can be written as
\begin{equation} \label{eq:tg}
{\rm prox}_{tg}(\bV) = \arg \min_{\bW \in \mathcal{W}} \frac{1}{2} \| \bV_{\rm sym} - \bW \|^2 - \frac{t}{\rho} \log \det(\bW)
\end{equation}
Let $\bV_{\rm sym} = \bU \bLam \bU^T$ be the symmetric eigendecomposition of $\bV_{\rm sym}$.
Also, let $\tilde{\bW} = \bU^T \bW \bU$, and note that $\bW \in \mathcal{W}$ implies $\tilde{\bW} \in \mathcal{W}$.
We have the following inequality for any $\bW \in \mathcal{W}$:
\begin{align}
\frac{1}{2} \| \bV_{\rm sym} - \bW \|^2 - \frac{t}{\rho} \log \det(\bW)
    & = \frac{1}{2} \| \bLam - \tilde{\bW} \|^2 - \frac{t}{\rho} \log \det(\tilde{\bW}) \nonumber \\
    & \geq \sum_{i=1}^n \frac{1}{2} ( \lambda_i - \tilde{w}_{ii} )^2 - \frac{t}{\rho} \log ( \tilde{w}_{ii} ) \nonumber \\
    & \geq \sum_{i=1}^n \min_{\tilde{w}_{ii} \geq \epsilon} \left[ \frac{1}{2} ( \lambda_i - \tilde{w}_{ii} )^2 - \frac{t}{\rho} \log ( \tilde{w}_{ii} ) \right]
    \label{eq:trt}
\end{align}
where the first equality is due to rotational invariance of the Euclidean norm and determinant;
the second inequality is due to $\| \bLam - \tilde{\bW} \|^2 \geq \sum_{i=1}^n ( \lambda_i - \tilde{w}_{ii} )^2$ and the Hadamard inequality $\det(\tilde{\bW}) \leq \prod_{i=1}^n \tilde{w}_{ii}$;
the third inequality is due to the fact that $\lambda_{\rm min}(\tilde{\bW}) \leq \tilde{w}_{ii}$ for all $i$.
One can readily show that the optimal solution to the problem in \eqref{eq:trt} is
$\tilde{w}_{ii}^\star = \max\left\{ \left(\lambda_i+\sqrt{\lambda_i^2+4t/\rho} \right)/2,\epsilon\right\}$.
Furthermore, by letting $\bW^\star = \bU \bD \bU^T$, $\bD = {\rm Diag}( \tilde{w}_{11}^\star,\ldots,\tilde{w}_{nn}^\star )$,
the equalities in \eqref{eq:trt} are attained.
Since $\bW^\star$ also lies in $\mathcal{W}$, we conclude that $\bW^\star$ is the optimal solution to the problem in \eqref{eq:tg}.

\subsection{Lipschitz Continuity of the Gradient of $f$} \label{app:Lip}

In this appendix we show that the function $f$ in \eqref{formulation} has a Lipschitz continuous gradient.
To this end, define $\bz=[(\text{vec}(\bW))^T,~\!\by^T]^T$ and
$$\phi_i(\bz)=\sqrt{\|\bC_i\bz\|^2+\epsilon}+\bd_i^T\bz-h_i, \quad i=1,\ldots,K,$$
where $\bC_i=[(\bg_i^T\otimes\bI),~\!\bm 0]$ (here ``$\otimes$'' denotes the Kronecker product) and  $\bd_i=[\bm 0^T,~\!\bg_i^T]^T$.
Then, $f$ can be written as $f(\bW,\by) = \sum_{i=1}^K \psi(\phi_i(\bz))$.
From the above equation, we see that $f$ has a Lipschitz continuous gradient if  every $\psi(\phi_i(\bz))$ has a Lipschitz continuous gradient.
Hence, we seek to prove the latter. Consider the following fact.
\begin{Fact}\label{fact:L-conti}
Let $\psi: \Rbb \rightarrow \Rbb$, $\phi: \Rbb^n \rightarrow \Rbb$ be functions that satisfy the following properties:
\begin{enumerate}[\leftmargin=1cm]
\item (a) $\psi'$ is bounded on $\Rbb$ and $\psi$ has a Lipschitz continuous gradient on $\Rbb$;


\item (b) $\nabla \phi$ is bounded on $\Rbb^n$ and $\phi$ has a Lipschitz continuous gradient on $\Rbb^n$.
\end{enumerate}
Then, $\psi(\phi(\bz))$ has a Lipschitz continuous gradient on $\Rbb^n$.
\end{Fact}
As Fact {\ref{fact:L-conti}} can be easily proved from the definition of Lipschitz continuity, its proof is omitted here for conciseness.
Recall that for our problem, $\psi$ is the one-sided Huber function.
One can verify that the one-sided Huber function has bounded $\psi'$ and Lipschitz continuous gradient.
As for $\phi_i$, let us first evaluate its gradient and Hessian
\begin{align*}
\nabla\phi_i(\bz) & = \frac{\bC_i^T\bC_i\bz}{\sqrt{\|\bC_i\bz\|^2+\epsilon}} + \bd_i, \\
\nabla^2 \phi_i(\bz) & = \frac{\bC_i^T\bC_i}{\sqrt{\|\bC_i\bz\|^2+\epsilon}}-\frac{(\bC_i^T\bC_i\bz) (\bC_i^T\bC_i\bz)^T}{(\|\bC_i\bz\|^2+\epsilon)^{3/2}}.
\end{align*}
We have
\begin{align*}
\|\nabla\phi_i(\bz)\|&\leq\|\bd_i\|+\frac{\|\bC_i^T\bC_i\bz\|}{\sqrt{\|\bC_i\bz\|^2+\epsilon}}\leq\|\bd_i\|+\frac{\sigma_\text{max}(\bC_i)\|\bC_i\bz\|}{\sqrt{\|\bC_i\bz\|^2+\epsilon}}\leq\|\bd_i\|+\sigma_\text{max}(\bC_i),
\end{align*}
where $\sigma_{\rm max}(\bX)$ denotes the largest singular value of $\bX$.
Hence, $\nabla\phi_i(\bz)$ is bounded.
Moreover, recall that a function has a Lipschitz continuous gradient if its Hessian is bounded.
Since
\begin{align*}
\| \nabla^2 \phi_i(\bz) \| &  \leq \sqrt{n+n^2} \lambda_{\rm max}( \nabla^2 \phi_i(\bz) )
 \leq \sqrt{n+n^2}  \lambda_{\rm max}\left(\frac{\bC_i^T\bC_i}{\sqrt{\|\bC_i\bz\|^2+\epsilon}} \right) \leq \frac{ \sqrt{n+n^2} \lambda_{\rm max} (\bC_i^T \bC_i )}{ \sqrt{\epsilon} },
\end{align*}
the function $\phi_i$ has a Lipschitz continuous gradient.
The desired result is therefore proven.

\bibliographystyle{siamplain}

\end{document}